\documentclass[runningheads]{llncs}

\usepackage{eccv}

\usepackage{eccvabbrv}
\usepackage{subcaption}
\usepackage{graphicx}
\usepackage{booktabs}

\usepackage[accsupp]{axessibility}  

\usepackage{longtable}
\usepackage{ragged2e} 
\usepackage{tabularx}
\usepackage{float}
\usepackage{makecell}
\usepackage{array}
\usepackage{placeins}
\usepackage[table]{xcolor}
\usepackage{booktabs}

\definecolor{headerblue}{RGB}{210,225,245}
\definecolor{rowgray}{RGB}{245,245,245}

\usepackage[pagebackref,breaklinks,colorlinks,citecolor=eccvblue]{hyperref}
\usepackage{orcidlink}
\usepackage[table]{xcolor}
\usepackage{colortbl}
\usepackage{booktabs}
\usepackage{subcaption}
\usepackage[table]{xcolor}
\usepackage{colortbl}
\usepackage{booktabs}
\usepackage{multirow}
\usepackage{booktabs}
\usepackage[table]{xcolor}
\usepackage{graphicx}
\usepackage{booktabs}
\usepackage[table]{xcolor}
\usepackage{graphicx}
\usepackage[most]{tcolorbox}
\definecolor{promptblue}{RGB}{220,235,255}
\definecolor{headerblue}{RGB}{210,225,245}
\definecolor{rowgray}{RGB}{248,249,252}
\definecolor{tableheader}{RGB}{210,225,245}
\definecolor{tablerow}{RGB}{245,248,252}
\usepackage{xcolor}
\definecolor{tableheader}{RGB}{45, 92, 160}
\definecolor{rowshade}{RGB}{235, 242, 252}
\definecolor{highlight}{RGB}{0,120,80}
\definecolor{tableheader}{RGB}{214,226,245}
\definecolor{rowshade}{RGB}{245,247,252}
\definecolor{tableheader}{RGB}{45, 92, 160}
\definecolor{rowshade}{RGB}{235, 242, 252}
\definecolor{highlight}{RGB}{0,120,80}
\begin{document}

\title{MedSPOT: A Workflow-Aware Sequential Grounding Benchmark for Clinical GUI} 

\titlerunning{MedSPOT}

\author{Rozain Shakeel\inst{1} \and
Abdul Rahman Mohammad Ali\inst{2} \and
Muneeb Mushtaq\inst{1} \and
Tausifa Jan Saleem\inst{3} \and 
Tajamul Ashraf\inst{1,4}}

\authorrunning{Shakeel et al.}

\institute{Gaash Research Lab, National Institute of Technology Srinagar, India \and
e\& Group, UAE \and
Mohammad Bin Zayed University of Artificial Intelligence (MBZUAI), UAE \and King Abdullah University of Science and Technology (KAUST), Saudi Arabia 
} 
\maketitle

\vspace{-0.8em}
\begin{center}
{\small Project Page: \url{https://rozainmalik.github.io/MedSPOT_web/}}
\end{center}
\vspace{-0.5em}

\begin{abstract}
  Despite the rapid progress of Multimodal Large Language Models (\texttt{MLLMs}), their ability to perform reliable visual grounding in high-stakes clinical software environments remains underexplored. Existing \texttt{GUI} benchmarks largely focus on isolated, single-step grounding queries, overlooking the sequential, workflow-driven reasoning required in real-world medical interfaces, where tasks evolve across interdependent steps and dynamic interface states. We introduce \textbf{MedSPOT}, a workflow-aware sequential grounding benchmark for clinical \texttt{GUI} environments. Unlike prior benchmarks that treat grounding as a standalone prediction task, \textbf{MedSPOT} models procedural interaction as a sequence of structured spatial decisions. The benchmark comprises 216 task-driven videos with 597 annotated keyframes, in which each task comprises 2--3 interdependent grounding steps within realistic medical workflows. This design captures interface hierarchies, contextual dependencies, and fine-grained spatial precision under evolving conditions. To evaluate procedural robustness, we propose a strict sequential evaluation protocol that terminates task assessment upon the first incorrect grounding prediction, explicitly measuring error propagation in multi-step workflows. We further introduce a comprehensive failure taxonomy, including edge bias, small-target errors, no prediction, near miss, far miss, and toolbar confusion, to enable systematic diagnosis of model behavior in clinical \texttt{GUI} settings. By shifting evaluation from isolated grounding to workflow-aware sequential reasoning, \textbf{MedSPOT} establishes a realistic and safety-critical benchmark for assessing multimodal models in medical software environments. 
  Code and data are available at \href{https://github.com/Tajamul21/MedSPOT}{https://github.com/Tajamul21/MedSPOT}
\end{abstract}

\section{Introduction}
\begin{table*}[t]
\centering
\rowcolors{2}{gray!6}{white}
\setlength{\tabcolsep}{8pt}
\renewcommand{\arraystretch}{1.2}
\caption{
Comparison of \textbf{MedSPOT} against existing \texttt{GUI} grounding benchmarks. 
Green cells indicate the presence of a capability, red cells indicate absence. 
\textbf{MedSPOT} is the only benchmark that jointly evaluates workflow-aware sequential grounding, early-termination protocol, structured failure taxonomy, and medical software environments.
}
\resizebox{\textwidth}{!}{
\begin{tabular}{lcccccc}
\toprule
\rowcolor{blue!12}
\textbf{Benchmark} & \textbf{Size} & \textbf{Multi-Step} & \textbf{Step Eval.} & \textbf{Sequential} & \textbf{Failure} & \textbf{Medical} \\
 &  & \textbf{Tasks} &  & \textbf{Protocol} & \textbf{Taxonomy} & \textbf{Software} \\
\midrule
ScreenSpot~\cite{cheng2024seeclick}       
& 1,272  
& \cellcolor{red!15}\texttimes 
& \cellcolor{red!15}\texttimes 
& \cellcolor{red!15}\texttimes 
& \cellcolor{red!15}\texttimes 
& \cellcolor{red!15}\texttimes \\

ScreenSpot-Pro~\cite{li2025screenspotpro} 
& 1,581  
& \cellcolor{red!15}\texttimes 
& \cellcolor{green!18}\checkmark 
& \cellcolor{red!15}\texttimes 
& \cellcolor{red!15}\texttimes 
& \cellcolor{red!15}\texttimes \\

OSWorld-G~\cite{xie2024osworld}           
& 564    
& \cellcolor{green!18}\checkmark 
& \cellcolor{red!15}\texttimes 
& \cellcolor{red!15}\texttimes 
& \cellcolor{red!15}\texttimes 
& \cellcolor{red!15}\texttimes \\

OmniACT~\cite{kapoor2024omniact}          
& 9,802  
& \cellcolor{green!18}\checkmark 
& \cellcolor{green!18}\checkmark 
& \cellcolor{red!15}\texttimes 
& \cellcolor{red!15}\texttimes 
& \cellcolor{red!15}\texttimes \\

AssistGUI~\cite{gao2024assistgui}         
& 100    
& \cellcolor{green!18}\checkmark 
& \cellcolor{green!18}\checkmark 
& \cellcolor{red!15}\texttimes 
& \cellcolor{red!15}\texttimes 
& \cellcolor{red!15}\texttimes \\

Mind2Web~\cite{deng2023mind2web}          
& 2,350  
& \cellcolor{green!18}\checkmark 
& \cellcolor{red!15}\texttimes 
& \cellcolor{red!15}\texttimes 
& \cellcolor{red!15}\texttimes 
& \cellcolor{red!15}\texttimes \\

\midrule
\rowcolor{blue!6}
\textbf{MedSPOT (Ours)} 
& \textbf{216 / 597} 
& \cellcolor{green!25}\checkmark 
& \cellcolor{green!25}\checkmark 
& \cellcolor{green!25}\checkmark 
& \cellcolor{green!25}\checkmark 
& \cellcolor{green!25}\checkmark \\
\bottomrule
\end{tabular}
}

\label{tab:benchmark_comparison}
\end{table*}
\begin{figure}[t]
    \centering
   \begin{subfigure}[t]{0.48\linewidth}
        \centering
        \includegraphics[width=\linewidth]{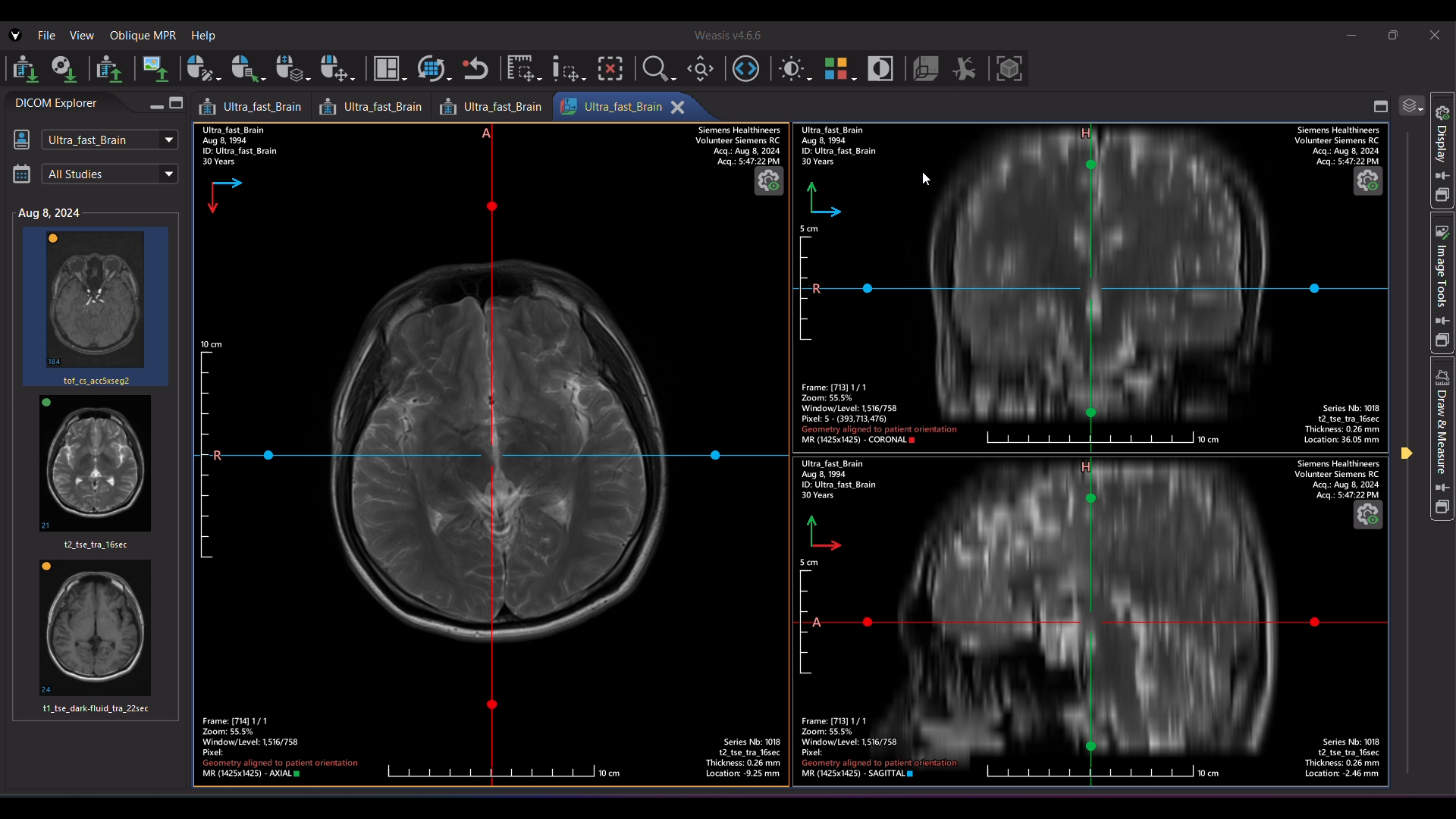}
        \caption{Medical \texttt{GUI} interface}
        \label{fig:medical_gui}
    \end{subfigure}
    \hfill
    \begin{subfigure}[t]{0.48\linewidth}
        \centering
        \includegraphics[width=\linewidth]{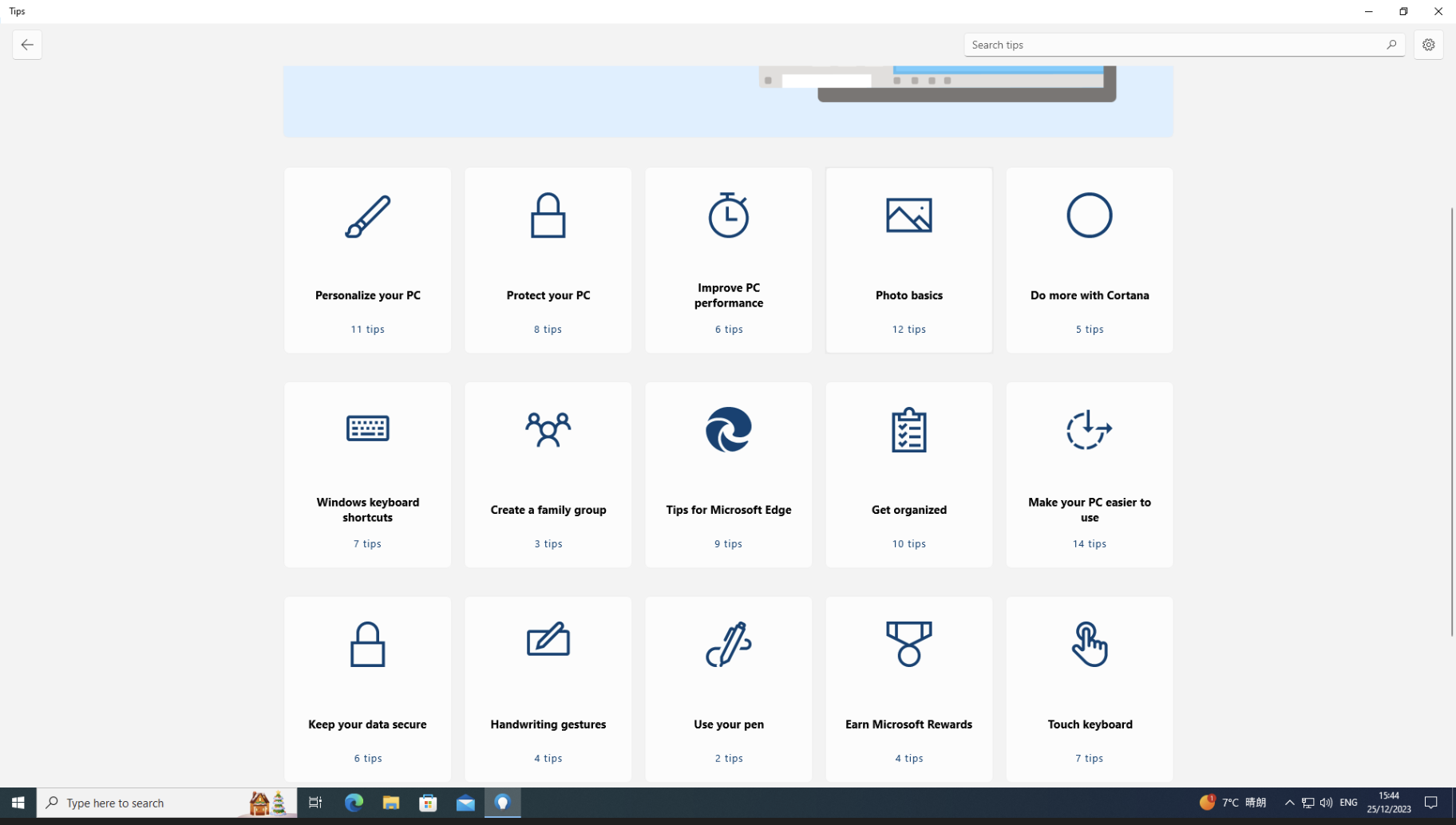}
        \caption{General \texttt{GUI} interface}
        \label{fig:simple_gui}
    \end{subfigure}
    \caption{Comparison of interface complexity. (a) Medical \texttt{GUI} environments are densely structured, hierarchically organized, and semantically rich with domain-specific terminology. (b) General \texttt{GUI} environments typically involve simpler layouts and short-horizon navigation tasks.}
    \label{fig:comparison}
\end{figure}

Graphical User Interfaces (\texttt{GUIs}) are the primary medium through which users interact with modern software systems. 
\textit{\texttt{GUI} grounding} refers to the task of mapping natural language instructions to precise visual and interactive elements within an interface~\cite{cheng2024seeclick, wu2025gui}. 
Recent advances in Vision-Language Models (\texttt{VLMs}) and Multimodal Large Language Models (\texttt{MLLMs})~\cite{qwen,singh2025openai,gpt4,qwen2.5-VL,qwen3vltechnicalreport} have significantly improved multimodal reasoning capabilities, enabling systems to interpret screenshots, detect interface elements, reason over contextual cues, and predict executable actions.

Despite this progress, existing research predominantly focuses on general-purpose environments such as web navigation and desktop automation. Benchmarks such as ScreenSpot~\cite{cheng2024seeclick}, Mind2Web~\cite{deng2023mind2web}, OmniAct~\cite{kapoor2024omniact}, and related works~\cite{li2025screenspotpro,zheng2024seeact,gou2024uground, ashraf2025agentx} evaluate grounding in short-horizon, single-step scenarios. These settings treat grounding as an isolated prediction problem, overlooking the structured, sequential reasoning required in real-world workflows.
This limitation becomes critical in high-stakes domains such as clinical software systems. Modern medical workflows rely heavily on complex graphical interfaces, including diagnostic imaging viewers, treatment planning systems, and health record dashboards. As illustrated in Fig.~\ref{fig:comparison}, medical \texttt{GUIs} are densely structured, hierarchically organized, and semantically rich with domain-specific terminology. Within such environments, accurate grounding requires not only visual alignment but also contextual awareness of interface hierarchies, multi-step procedural reasoning, and strict precision. An early grounding error can invalidate subsequent actions, potentially leading to severe operational consequences. 

Yet, despite the rapid growth of both \texttt{GUI} grounding and medical \texttt{AI} research, there exists no benchmark specifically designed to evaluate \textit{workflow-aware sequential grounding} in clinical software environments. Existing \texttt{GUI} datasets focus on general-purpose settings, while medical benchmarks~\cite{yue2025medsgbench,chen2024gmaimmbench,Hu2024OmniMedVQAAN} primarily evaluate static image reasoning rather than interactive procedural workflows. As a result, the intersection of sequential grounding and safety-critical medical interfaces remains largely unexplored.  As summarized in Table~\ref{tab:benchmark_comparison}, existing \texttt{GUI} grounding benchmarks primarily focus on general-purpose environments and short-horizon tasks. While several datasets support multi-step interactions, none incorporate a strict sequential evaluation protocol, structured failure analysis, or domain-specific clinical software settings. In contrast, \textbf{MedSPOT} is the first benchmark to jointly evaluate workflow-aware grounding under early-termination scoring in safety-critical medical interfaces.

\begin{figure}[t]
    \centering   
    \includegraphics[width=1\linewidth]{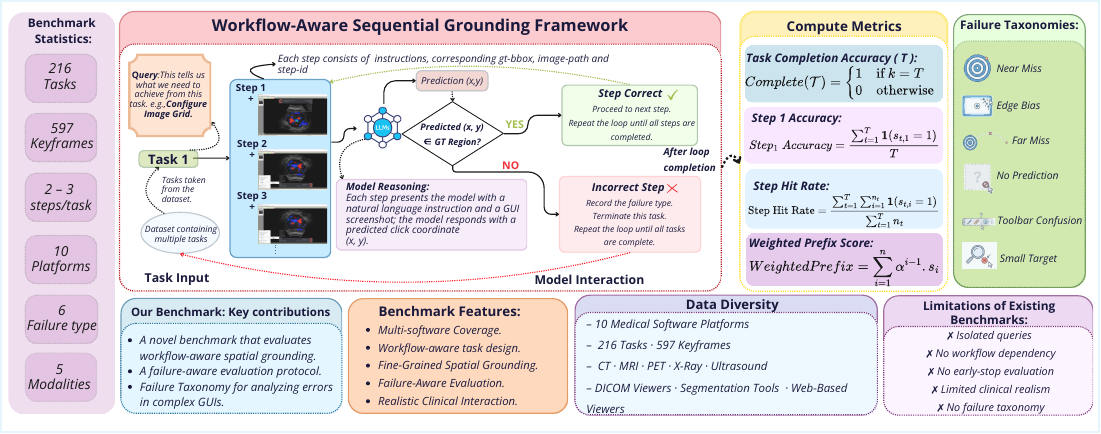}
    \caption{Overview of the \textbf{MedSPOT} dataset construction and sequential evaluation pipeline.}
    \label{fig:pipeline}
    \vspace{-0.5 cm}
\end{figure}

To address this gap, we introduce \textbf{MedSPOT}, a benchmark for evaluating workflow-aware spatial grounding in clinical \texttt{GUI} environments. \textbf{MedSPOT} spans \textbf{10 functionally diverse medical imaging software platforms} and consists of \textbf{216 task-driven videos with 597 annotated keyframes}, as illustrated in Fig.~\ref{fig:pipeline}. Each task involves multiple interdependent grounding steps that reflect realistic clinical procedures, where decisions evolve across dynamic interface states.
Beyond dataset construction, we propose a \textit{failure-aware sequential evaluation protocol} that terminates task scoring upon the first incorrect grounding prediction. This strict early-termination strategy explicitly measures error propagation across multi-step workflows, providing a more realistic assessment of procedural robustness. Furthermore, we introduce a structured six-class failure taxonomy, including edge bias, small-target errors, no prediction, near miss, far miss, and toolbar confusion, to systematically diagnose grounding behavior in complex medical interfaces. \textcolor{gray}{More details regarding MedSPOT are given in Supplementary Material~\ref{datasetcard}.}

\noindent\textbf{Our contributions are summarized as follows:}
\begin{itemize}
    \item We introduce \textbf{MedSPOT}, the first benchmark for workflow-aware sequential grounding in clinical \texttt{GUI} environments, spanning 10 medical software platforms with 216 tasks and 597 annotated keyframes.
    \item We propose a \textit{failure-aware sequential evaluation protocol} with early termination, capturing realistic error propagation in safety-critical multi-step workflows.
    \item We define a structured \textit{six-class failure taxonomy} to systematically analyze grounding errors in densely structured medical interfaces.
    \item We conduct a comprehensive evaluation of 16 state-of-the-art \texttt{MLLMs}, revealing a substantial performance gap between general-purpose models and \texttt{GUI}-specialized architectures in clinical environments.
\end{itemize}
\section{Related Work}

\subsection{GUI Grounding Benchmarks and Agent Models}

Recent benchmarks have advanced \texttt{GUI} grounding in general-purpose environments. ScreenSpot~\cite{cheng2024seeclick} introduced a cross-platform grounding dataset across mobile, desktop, and web interfaces, while ScreenSpot-Pro~\cite{li2025screenspotpro} extended evaluation to high-resolution professional applications, highlighting degradation on small, dense \texttt{UI} elements. VenusBench-GD~\cite{zhou2025venusbenchgd} and MMBench-GUI~\cite{wang2025mmbenchgui} further expanded coverage with hierarchical evaluation protocols, and web-centric datasets such as Mind2Web~\cite{deng2023mind2web} and OmniACT~\cite{kapoor2024omniact} incorporated multi-step browsing tasks.
Parallel to benchmark development, \texttt{GUI} grounding models have progressed rapidly. UI-TARS~\cite{qin2025ui} integrates perception, unified action modeling, and System-2 reasoning at scale; Agent-X~\cite{ashraf2025agentx} emphasizes structured multi-step reasoning; and GUI-Actor~\cite{wu2025gui} proposes coordinate-free grounding via a dedicated \texttt{<ACTOR>} token. General-purpose backbones such as Qwen2.5-VL~\cite{qwen2.5-VL}, Qwen3-VL~\cite{qwen3vltechnicalreport}, and GPT-style models remain strong baselines.
Despite this progress, existing benchmarks focus on general-purpose software, evaluate largely independent steps without strict inter-step dependency, and lack early-termination protocols to measure error propagation. As summarized in Table~\ref{tab:benchmark_comparison}, none jointly address sequential dependency, structured failure analysis, and clinical software environments. \textbf{MedSPOT} fills this gap by modeling grounding as interdependent spatial decisions under a strict sequential evaluation protocol. Desktop GUI grounding has been explored in~\cite{ettifouri2024visual, qian2024visual}. Concurrent work Chain-of-Ground~\cite{li2025chainofgroundimprovingguigrounding} improves single-step grounding via iterative reasoning, unlike MedSPOT's multi-step sequential evaluation.

\subsection{Medical Benchmarks and Clinical Software Environments}

Medical AI benchmarks have predominantly focused on static image understanding and clinical question answering~\cite{dmaster}. Datasets such as MedSG-Bench~\cite{yue2025medsgbench}, GMAI-MMBench~\cite{chen2024gmaimmbench}, and OmniMedVQA~\cite{Hu2024OmniMedVQAAN} evaluate multimodal reasoning over radiological images and medical reports. VividMed~\cite{luo2025vividmed} advances medical visual grounding but operates on static images. While these works advance domain-specific reasoning, they do not consider interactive \texttt{GUI}-based workflows or spatial grounding within software interfaces.
In practice, modern clinical workflows rely heavily on specialized software platforms including Orthanc~\cite{Jodogne2018}, Weasis~\cite{weasis}, RadiAnt~\cite{radiant}, BlueLight DICOM Viewer~\cite{chen2023bluelight}, MicroDICOM~\cite{MicroDicom}, 3D Slicer~\cite{slicer2015}, ITK-SNAP~\cite{py06nimg}, MITK~\cite{wolf2005}, Ginkgo CADx~\cite{ginkgocadx}, and DICOMscope~\cite{dicomscope}. These systems feature densely structured toolbars, hierarchical menus, domain-specific terminology, and fine-grained interaction mechanisms that impose unique grounding challenges not captured in existing GUI datasets.
To our knowledge, \textbf{MedSPOT} is the first benchmark to systematically evaluate workflow-aware sequential grounding within real-world clinical \texttt{GUI} environments, bridging the gap between multimodal reasoning research and safety-critical medical software interaction.

\section{MedSPOT Benchmark and Evaluation}
\label{sec:Benchmark}

\textbf{Problem Formulation and Dataset Design.}
\label{subsec:dataset_design}
We formulate workflow-aware \texttt{GUI} grounding in clinical software as a \textit{sequential, instruction-conditioned spatial localization problem}. Unlike conventional grounding benchmarks that evaluate isolated predictions, \textbf{MedSPOT} models grounding as a temporally dependent sequence of spatial decisions within evolving interface states.

Let $\mathcal{S}$ denote the set of medical software platforms. The dataset is defined as: $\mathcal{D} = \bigcup_{s \in \mathcal{S}} \mathcal{D}_s,$ where $\mathcal{D}_s$ represents the task set associated with software $s$.
Each $\mathcal{D}_s$ consists of $N_s$ task workflows:
$\mathcal{D}_s = \{ \mathcal{T}_i \}_{i=1}^{N_s}.$
Each task $\mathcal{T}$ is a finite ordered sequence of $T$ interaction steps:
$\mathcal{T} = \{ (I_t, s_t, A_t) \}_{t=1}^{T},$ where:
\begin{itemize}
    \item $I_t \in \mathbb{R}^{H \times W \times 3}$ denotes the \texttt{GUI} frame at step $t$,
    \item $s_t$ is the natural language instruction,
    \item $A_t$ is the corresponding ground-truth action.
\end{itemize}

Each action is defined as: $A_t = (a_t, y_t, B_t),$ where:
\begin{itemize}
    \item $a_t \in \mathcal{A} = \{\texttt{click}\}$,
    \item $y_t$ is the semantic target description,
    \item $B_t = (x, y, w, h) \in [0,100]^4$ is the normalized bounding box in percentage coordinates.
\end{itemize}

Importantly, the steps within $\mathcal{T}$ are \textit{interdependent}. An incorrect grounding at time $t$ invalidates downstream actions, reflecting the procedural dependency structure inherent in clinical workflows. This sequential dependency differentiates \textbf{MedSPOT} from prior \texttt{GUI} grounding datasets that treat each instruction-frame pair independently.

\subsection{Annotation Protocol}

\begin{figure}[t]
    \centering   
    \includegraphics[width=1\linewidth]{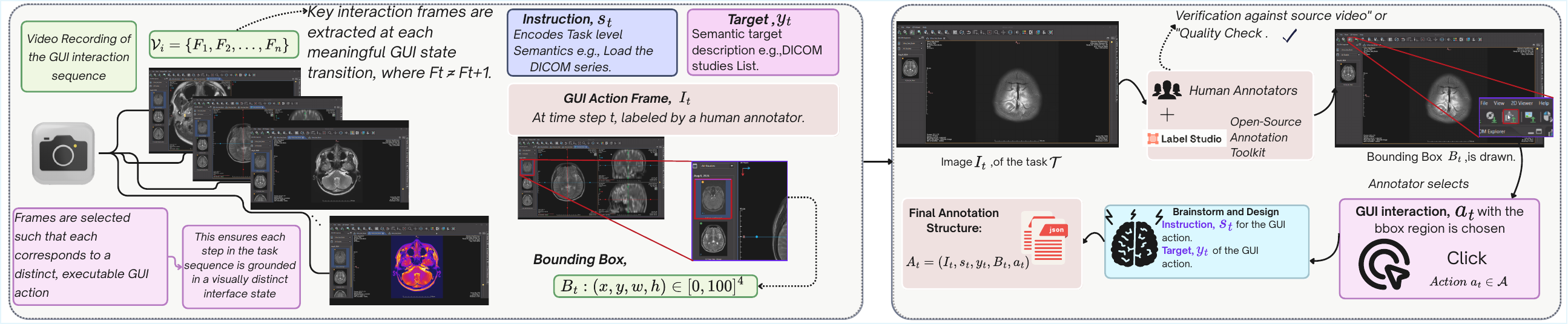}
    \caption{Overview of the \textbf{MedSPOT} annotation pipeline. \texttt{GUI} interaction videos are recorded from real clinical software, segmented into causally consistent decision frames, and annotated with step-level instructions, semantic targets, bounding boxes, and action types to construct sequential grounding tasks.}
    \label{fig:annotation_pipeline}
\end{figure}

For each medical software platform $s \in \mathcal{S}$, we record real \texttt{GUI} interaction workflows as video sequences:
$\mathcal{V}_i = \{ F_1, F_2, \dots, F_n \}, \quad n \geq 2,$
where each action induces a visible \texttt{GUI} state transition:
$F_t \neq F_{t+1}.$ These sequences simulate realistic clinical workflows, including loading DICOM studies, navigating image series, applying transformations, performing measurements, and exporting results. As illustrated in Fig.~\ref{fig:annotation_pipeline}, the annotation process converts raw clinical \texttt{GUI} interaction videos into temporally ordered, step-level grounding tasks through structured segmentation and multi-level labeling.

\paragraph{Step Extraction.}
From each video $\mathcal{V}_i$, we extract a minimal ordered subset of decision frames:
$\{ I_t \}_{t=1}^{T} \subset \mathcal{V}_i,$ where each $I_t$ corresponds to a meaningful interaction decision point and preserves temporal order. This produces a compact sequence capturing causally consistent state transitions.

\paragraph{Frame-Level Annotation.}
Each frame is annotated using Label Studio~\cite{LabelStudio} as:
$\mathcal{A}_t = (I_t, s_t, y_t, B_t, a_t),$
where:

\begin{itemize}
    \item $I_t$ is the \texttt{GUI} frame encoding layout structure, widgets, modality views, and domain-specific elements,
    \item $s_t$ is the natural language instruction,
    \item $y_t$ is the semantic description of the target \texttt{UI} element,
    \item $B_t = (x,y,w,h) \in [0,100]^4$ is the normalized bounding box,
    \item $a_t \in \mathcal{A} = \{\texttt{click}\}$.
\end{itemize}
 
This annotation enforces joint learning of:
$
P(\text{region} \mid s_t, I_t)
\quad \text{and} \quad
P(y_t \mid I_t),
$
linking semantic reasoning and spatial localization.
\textcolor{gray}{Inter-annotator agreement details are reported in the Supplementary material~\ref{sec:iaa}.}
\paragraph{Task Construction.}
Each annotated interaction step is represented as $\text{step}_t = (\text{id}_t, I_t, s_t, A_t)$, where $\text{id}_t$ denotes the step index, $I_t$ is the \texttt{GUI} frame, $s_t$ is the natural language instruction, and $A_t$ is the corresponding ground-truth action. The steps are ordered temporally as $\text{step}_1 < \text{step}_2 < \dots < \text{step}_T$, forming a causally consistent interaction trajectory in which the next interface state depends on the previous state and executed action, i.e., $P(I_{t+1} \mid I_t, a_t)$. Each recorded \texttt{GUI} interaction video produces a task $\mathcal{T}_i = \{\text{step}_1, \dots, \text{step}_T\}$, representing a complete procedural workflow. Tasks are grouped per software platform as $\mathcal{D}_s = \{\mathcal{T}_1, \dots, \mathcal{T}_{N_s}\}$, with disjoint task sets such that $\mathcal{T}_i \cap \mathcal{T}_j = \emptyset$ for $i \neq j$. This hierarchical construction (step $\rightarrow$ task $\rightarrow$ software $\rightarrow$ dataset) ensures completeness, non-overlapping workflows, and preservation of causal structure. Overall, \textbf{MedSPOT} jointly evaluates spatial grounding and procedural reasoning within structured medical \texttt{GUI} workflows.

\subsection{Evaluation Pipeline}
\label{subsec:Evaluation_protocols}

Given a \texttt{GUI} frame $I_t$, instruction $s_t$, and interaction history $\mathcal{H}_t$, a model predicts click coordinates $\hat{p}_t \sim P(p \mid I_t, s_t, \mathcal{H}_t)$. Models may output a single point or a set of top-$K$ candidates $\hat{P}_t = \{\hat{p}_{t,k}\}_{k=1}^K$, with $\hat{p}_{t,k} \in \mathbb{R}^2$. Ground-truth annotations are provided as normalized bounding boxes $(x,y,w,h) \in [0,100]^4$, which are converted into pixel space as $B_t = (x_1,y_1,x_2,y_2)$ using the image dimensions $(W,H)$. A prediction is considered correct if any predicted point lies within $B_t$, i.e., $x_1 \le \hat{x} \le x_2$ and $y_1 \le \hat{y} \le y_2$. For models operating on discrete visual patches, a tolerance region $\hat{B}_p = (\hat{x} \pm \delta, \hat{y} \pm \delta)$ is introduced, and a hit is accepted if $\hat{B}_p \cap B_t \neq \emptyset$.

To analyze spatial difficulty, we define the normalized target area $A(B) = \frac{(x_2-x_1)(y_2-y_1)}{WH}$ and categorize targets as small, medium, or large based on predefined thresholds, capturing the nonlinear degradation observed for compact \texttt{UI} elements~\cite{zhao2025guicursor}. 

Evaluation follows a strict sequential protocol. Step correctness is defined as $\delta_t = 1$ if $\hat{p}_t \in B_t$ and $0$ otherwise. We compute the longest valid prefix $k = \max \{ t \mid \forall i \le t, \delta_i = 1 \}$ and terminate evaluation upon the first failure. A task is considered complete only if $k = T$, meaning all steps are correct. This corresponds to maximizing the joint probability $\prod_{t=1}^{T} P(\hat{p}_t \in B_t \mid \mathcal{H}_t)$, implying that if per-step accuracy is $p$, the probability of full task completion scales as $p^T$. Consequently, the protocol emphasizes temporal consistency and penalizes early errors, transforming evaluation from independent step accuracy into a measure of causally consistent, workflow-aware grounding. \textcolor{gray}{Further details are in the Supplementary material~\ref{sec:stats}.}

\subsection{Domain Coverage and Interface Diversity}

\textbf{MedSPOT} spans 10 open-source medical imaging platforms, covering three primary interface categories: (1) \texttt{DICOM/PACS} viewers, (2) segmentation and research tools, and (3) web-based viewers. The benchmark comprises \textbf{216} video tasks, \textbf{597} annotated keyframes, with an average of \textit{2--3 interdependent steps per task}.
The dataset captures heterogeneous \texttt{GUI} layouts, modality-specific visualization structures, hierarchical toolbars, nested menus, and diverse interaction paradigms. This diversity ensures evaluation of cross-interface generalization rather than overfitting to a single visual layout.

\noindent\textbf{Software Coverage.}
The platforms include BlueLight DICOM Viewer~\cite{chen2023bluelight}, 3D Slicer~\cite{slicer2015}, Ginkgo-CADx~\cite{ginkgocadx}, DICOMscope~\cite{dicomscope}, Orthanc~\cite{Jodogne2018}, RadiAnt~\cite{radiant}, MicroDICOM~\cite{MicroDicom}, Weasis~\cite{weasis}, MITK~\cite{wolf2005}, and ITK-SNAP~\cite{py06nimg}. These systems support diverse imaging modalities, including \texttt{CT}, \texttt{MRI}, \texttt{PET}, X-ray, and Ultrasound.

\begin{figure}[t]
    \centering

    \begin{subfigure}[t]{0.3\linewidth}
        \centering
        \includegraphics[width=\linewidth, height=4.5cm]{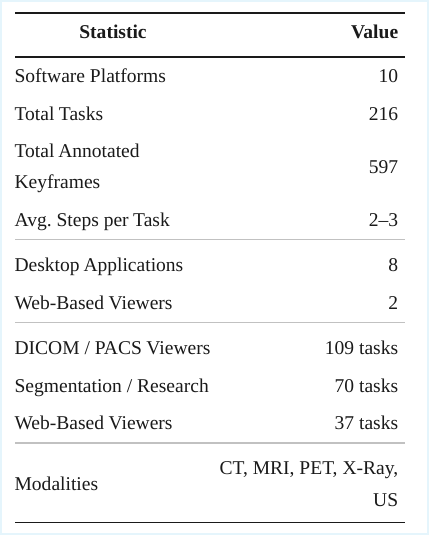}
        \caption{Dataset statistics}
        \label{fig:dataset_stats}
    \end{subfigure}
    \hfill
    \begin{subfigure}[t]{0.32\linewidth}
        \centering
        \includegraphics[width=\linewidth, height=4.5cm]{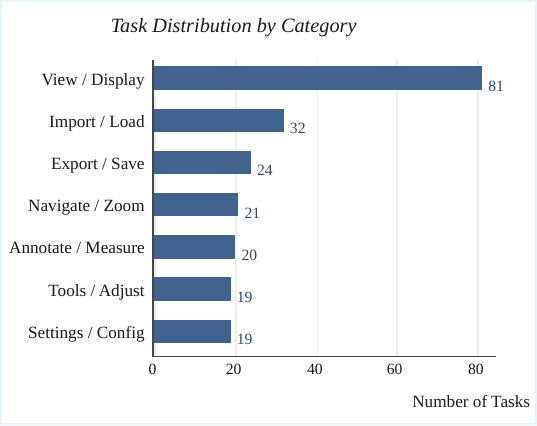}
        \caption{Task categories}
        \label{fig:task_categories}
    \end{subfigure}
    \hfill
    \begin{subfigure}[t]{0.32\linewidth}
        \centering
        \includegraphics[width=\linewidth, height=4.5cm]{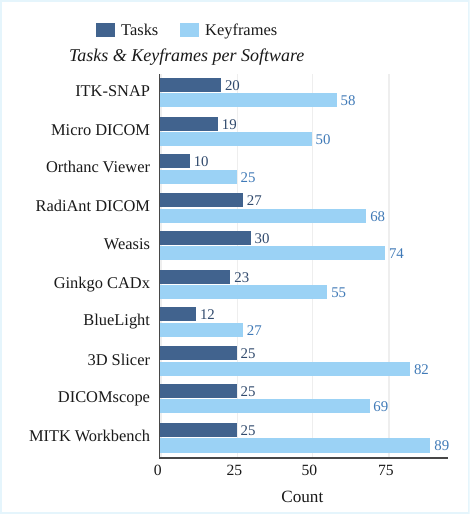}
        \caption{Per-software task depth}
        \label{fig:task_depth}
    \end{subfigure}

\caption{
\textbf{MedSPOT} dataset statistics illustrating overall scale, task category distribution, and per-software coverage across ten clinical platforms.
}
    \label{fig:dataset_overview}
\end{figure}
\noindent\textbf{Task Category Distribution.}
The overall dataset composition is summarized in Fig.~\ref{fig:dataset_overview}. 
Specifically, Fig.~\ref{fig:dataset_stats} reports the global benchmark statistics, including the total number of video tasks, annotated keyframes, and average steps per task. 
Figure~\ref{fig:task_categories} illustrates the distribution of tasks across seven functional categories. View/Display operations dominate (81 tasks), reflecting the central role of visualization in clinical imaging workflows. Additional categories include Import/Load (32), Export/Save (24), Navigate/Zoom (21), Annotate/Measure (20), Tools/Adjust (19), and Settings/Configuration (19), collectively covering the procedural spectrum encountered in real clinical usage. 
Finally, Fig.~\ref{fig:task_depth} presents the per-software task and keyframe distribution, highlighting variations in interaction depth and interface complexity across platforms. \textcolor{gray}{Details regarding data provenance and ethics are mentioned in the supplementary~\ref{sec:ethics}.}

\section{Experiments}

\textbf{Evaluation Protocol}
We employ two complementary evaluation functions:
\textit{Step-wise Evaluation.}
At each step $t$, a prediction is considered correct if the predicted click location lies within (or sufficiently overlaps) the ground-truth bounding box. Evaluation follows an early-termination strategy: once a step is incorrect, all subsequent steps in the task are invalidated. This mirrors real-world \texttt{GUI} usage, where an incorrect interaction alters the interface state and disrupts the intended procedural trajectory.
\textit{Task Completion.}
A task is considered successfully completed only if all interdependent steps are executed correctly. This metric provides a strict measure of procedural robustness and long-horizon reasoning capability. 
\begin{table*}[t]
\centering
\setlength{\tabcolsep}{8pt}
\renewcommand{\arraystretch}{1.15}
\caption{
Comparison of state-of-the-art \texttt{MLLMs} on \textbf{MedSPOT-Bench} under the strict sequential evaluation protocol. 
We report Step Hit Rate (\texttt{SHR}), Step-1 Accuracy (\texttt{S1A}), Weighted Prefix Score (\texttt{WPS}), and Task Completion Accuracy (\texttt{TCA}), where higher values indicate stronger spatial precision and sustained workflow consistency. 
\textbf{\color{green!60!black}Best results} are highlighted in bold green and \underline{\color{blue!70!black}second-best results} are underlined in blue.
}
\resizebox{\textwidth}{!}{%
\begin{tabular}{lccccc}
\toprule
\rowcolor{blue!12}
\textbf{Model Name} & \textbf{Params} & \textbf{SHR(\%)$\uparrow$} & \textbf{S1A(\%)$\uparrow$} & \textbf{WPS$\uparrow$} & \textbf{TCA(\%)$\uparrow$} \\
\midrule
\rowcolor{gray!15}
\multicolumn{6}{l}{\textbf{Closed Source Models}} \\
GPT 4o mini~\cite{openai2024gpt4omini}& - & 4.91 & 5.14 & 0.051 & 0.0 \\
GPT 5~\cite{singh2025openai} & - & 16.4 & 16.5 & 0.185 & 2.8 \\

\midrule
\rowcolor{gray!15}
\multicolumn{6}{l}{\textbf{Open Source Models}} \\

Llama 3.2 Vision-Instruct~\cite{llama_3_2} & 11B & 0.0 & 0.0 & 0.0 & 0.0 \\
Qwen2-VL-Instruct~\cite{Qwen2-VL} & 7B & 1.83 & 1.87 & 0.02 & 0.0 \\
DeepSeek-VL2~\cite{deepseek_vl2} & 16B & 2.7 & 2.8 & 0.03 & 0.0 \\
Gemma 3~\cite{gemmateam2025gemma3technicalreport} & 27B & 3.7 & 1.3 & 0.04 & 0.0 \\
Mistral-3-Instruct~\cite{liu2026ministral} & 8B & 6.14 & 6.0 & 0.064 & 0.0 \\
UGround-V1~\cite{gou2024uground} & 7B & 10.17 & 8.88 & 0.107 & 0.93 \\
Seeclick~\cite{cheng2024seeclick} & - & 4.3 & 9.3 & 0.11 & 1.4 \\
Qwen2.5-VL-Instruct~\cite{qwen2.5-VL} & 7B & 19.3 & 33.17 & 0.48 & 12.6 \\
CogAgent~\cite{hong2024cogagent} & 9B & 37.8 & 27.1 & 0.45 & 15.4 \\
Aguvis~\cite{xu2025aguvis} & 7B & 54.8 & 44 & 0.75 & 26.7 \\
Os-Atlas~\cite{wu2024atlas} & 7B & \underline{\color{blue!70!black}55} & 45 & 0.8 & 26.6 \\
UI-Tars 1.5-VL~\cite{qin2025ui} & 7B & \textbf{\color{green!60!black}66} & 57.5 & 1.0 & 30.8 \\
Qwen3-VL-Instruct~\cite{qwen3vltechnicalreport} & 8B & 46.6 & \underline{\color{blue!70!black}63.0} & \underline{\color{blue!70!black}1.1} & \underline{\color{blue!70!black}35.0} \\
GUI-Actor~\cite{wu2025gui} & 7B & 49.6 & \textbf{\color{green!60!black}65.0} & \textbf{\color{green!60!black}1.2} & \textbf{\color{green!60!black}43.5} \\

\bottomrule
\end{tabular}%
}
\label{tab:comparison}
\end{table*}

\medskip
\noindent\textbf{Metrics}
We report the following metrics to capture both spatial precision and sequential consistency:
\textit{Task Completion Accuracy (\texttt{TCA}).}
The fraction of tasks for which all steps are correctly executed. This is the most stringent measure of workflow reliability.
\textit{Step-1 Accuracy (\texttt{S1A}).}
The fraction of tasks in which the first step is correctly grounded. This isolates initial perception and instruction alignment without sequential compounding effects.
\textit{Step Hit Rate (\texttt{SHR}).}
The ratio of correctly completed steps before the first failure to the total number of evaluable steps across all tasks. This metric quantifies sequential progress under early stopping.
\textit{Weighted Prefix Score (\texttt{WPS}).}
To emphasize the importance of early correctness, we introduce a weighted prefix formulation:
$
\text{\texttt{WPS}} = \sum_{i=1}^{n} \alpha^{i-1} s_i,
$
where $s_i = 1$ if step $i$ is correct and $0$ otherwise, and $\alpha = 0.8$ is an exponential decay factor. This score reflects the disproportionate impact of early errors on overall workflow reliability. \textcolor{gray}{Further details regarding Inference Protocol are in the Supplementary material~\ref{sec:inferencep}.}

\medskip
\noindent\textbf{Failure Analysis.}
For each incorrect step, we record the failure category, \textcolor{gray}{according to the taxonomy described in the Supplementary material~\ref{sec:failure}}. This enables fine-grained diagnosis of systematic weaknesses, including spatial precision errors, small-target failures, toolbar confusion, and semantic misalignment.

\subsection{Benchmark Results}

\noindent\textbf{1. Large-Scale Comparison Across MLLMs.}
We conduct a comprehensive evaluation of state-of-the-art multimodal large language models (\texttt{MLLMs}) on MedSPOT using our strict failure-aware sequential protocol. As shown in Table~\ref{tab:comparison}, performance is reported across both step-level and task-level metrics, including Step Hit Rate (\texttt{SHR}), Step-1 Accuracy (\texttt{S1A}), Weighted Prefix Score (\texttt{WPS}), and Task Completion Accuracy (\texttt{TCA}).
The results reveal a substantial performance gap between general-purpose vision-language models and interface-specialized architectures. Despite strong performance on conventional multimodal benchmarks, widely adopted general-purpose models struggle dramatically in workflow-aware medical \texttt{GUI} grounding.

\medskip
\noindent\textbf{2. Collapse of General-Purpose MLLMs.}
We observe near-complete failure among several widely used multimodal models. Llama 3.2 Vision-Instruct~\cite{llama_3_2}, Qwen2-VL-Instruct~\cite{Qwen2-VL}, DeepSeek-VL2~\cite{deepseek_vl2}, and Gemma 3~\cite{gemmateam2025gemma3technicalreport} all achieve 0\% Task Completion Accuracy (\texttt{TCA}), despite parameter sizes ranging from 7B to 27B. Their Step Hit Rate (\texttt{SHR}) remains below 6\%, and \texttt{WPS} is effectively negligible.
Closed-source frontier models exhibit similar fragility. GPT-4o-mini~\cite{openai2024gpt4omini} achieves only 4.91\% \texttt{SHR} with 0\% \texttt{TCA}, while GPT-5~\cite{singh2025openai} reaches 16.5\% \texttt{SHR} but only 2.8\% \texttt{TCA}. Although GPT-5 occasionally predicts the correct first step, it fails to sustain consistent multi-step execution under strict early termination.
These findings indicate that parameter scale and generic multimodal pretraining do not translate into reliable spatial grounding within structured clinical interfaces. Unlike natural-image reasoning tasks, medical \texttt{GUI} environments demand pixel-level precision, fine-grained discrimination among visually similar toolbar elements, and hierarchical interface understanding.

\begin{figure}[t]
    \centering
    \includegraphics[width=\linewidth]{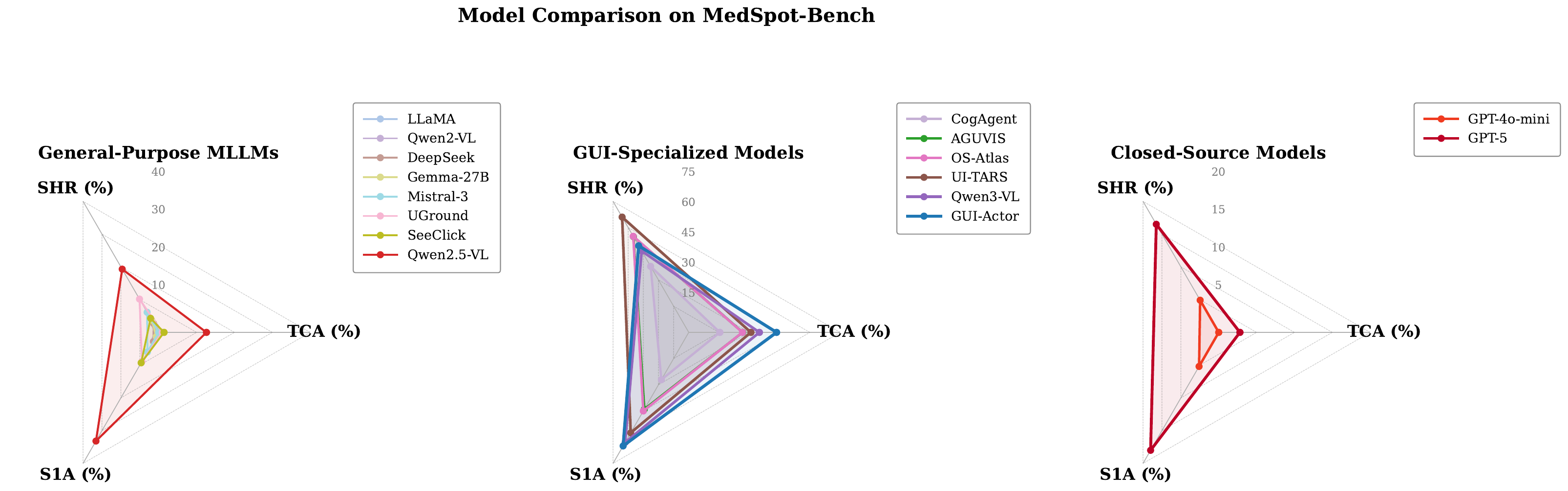}
    \caption{Comparison of representative models across Step-1 Accuracy (\texttt{S1A}), Step Hit Rate (\texttt{SHR}), and Task Completion Accuracy (\texttt{TCA}). The consistent gap between \texttt{S1A} and \texttt{TCA} highlights compounding sequential errors under strict workflow constraints.}
    \label{fig:modelcomparison}
\end{figure}

\medskip
\noindent\textbf{3. Sequential Fragility and Error Propagation.}
Even stronger multimodal models that demonstrate reasonable single-step performance exhibit substantial degradation under sequential evaluation. For example, Qwen3-VL-Instruct~\cite{qwen3vltechnicalreport} achieves 63.0\% Step-1 Accuracy (\texttt{S1A}), yet \texttt{TCA} drops to 35.0\%. GUI-Actor~\cite{wu2025gui} achieves the highest \texttt{S1A} (65.0\%) but completes only 43.5\% of tasks. UI-Tars 1.5-VL~\cite{qin2025ui} attains the highest \texttt{SHR} (66\%) while \texttt{TCA} remains limited to 30.8\%.
This consistent drop from \texttt{S1A} to \texttt{SHR} to \texttt{TCA} reflects compounding error under our prefix-constrained evaluation. As visualized in Fig.~\ref{fig:modelcomparison}, the performance hierarchy consistently follows:
$
\text{\texttt{TCA}} \ll \text{\texttt{SHR}} < \text{\texttt{S1A}}.
$
Even moderate per-step error rates rapidly reduce full task completion probability due to early termination and state dependency.

\medskip
\noindent\textbf{4. Interface-Specialized Architectures.}
Models explicitly designed for \texttt{GUI} interaction perform substantially better. Aguvis~\cite{xu2025aguvis} and Os-Atlas~\cite{wu2024atlas} achieve approximately 55\% \texttt{SHR} and 26\% \texttt{TCA}. CogAgent~\cite{hong2024cogagent} achieves 37.8\% \texttt{SHR} but only 15.4\% \texttt{TCA}, further demonstrating degradation under multi-step constraints.
The strongest overall model, GUI-Actor~\cite{wu2025gui}, achieves 43.5\% \texttt{TCA} (Table~\ref{tab:comparison}). Nevertheless, even this model fails to complete more than half of the tasks, indicating that workflow-aware reasoning in high-stakes medical \texttt{GUIs} remains an open challenge.

\medskip
\noindent\textbf{5. Weighted Prefix Score and Sustained Consistency.}
To quantify partial sequential robustness, we analyze \texttt{WPS}. As shown in Table~\ref{tab:comparison}, \texttt{WPS} strongly correlates with \texttt{TCA}. GUI-Actor achieves the highest \texttt{WPS} (1.2) alongside the highest \texttt{TCA} (43.5\%), followed by Qwen3-VL-Instruct (1.1 \texttt{WPS}, 35.0\% \texttt{TCA}) and UI-Tars (1.0 \texttt{WPS}, 30.8\% \texttt{TCA}). In contrast, general-purpose \texttt{MLLMs} exhibit near-zero \texttt{WPS}, reflecting early-stage failure and minimal sustained prefix correctness.
Overall, \textbf{MedSPOT} exposes a fundamental limitation of current multimodal systems: accurate first-step grounding does not imply reliable workflow execution. Robust medical \texttt{GUI} interaction requires sustained spatial precision and temporally consistent reasoning across interdependent steps.

\begin{table*}[t]
\centering
\setlength{\tabcolsep}{6pt}
\renewcommand{\arraystretch}{1.15}
\caption{
\textbf{Failure taxonomy distribution on \textbf{MedSPOT-Bench}.}
Open-source models are shaded in blue and closed-source models in red. 
Worst failure counts are highlighted in \textbf{\color{red!70!black}bold red}, while lowest values are \underline{\color{blue!70!black}underlined blue}.
}
\resizebox{\textwidth}{!}{
\begin{tabular}{lccccccc}
\toprule
\rowcolor{blue!20}
\textbf{Model Name} & \textbf{Params} & 
\textbf{Edge Bias} & 
\textbf{Far Miss} & 
\textbf{Toolbar Confusion} & 
\textbf{Near Miss} & 
\textbf{No Prediction} & 
\textbf{Small Target} \\
\midrule

\rowcolor{blue!8}
\multicolumn{8}{l}{\textbf{Open Source Models}} \\

\rowcolor{blue!5}
Llama 3.2 Vision-Instruct~\cite{llama_3_2} & 11B  & 49 & \underline{\color{blue!70!black}7} & 4 & \underline{\color{blue!70!black}0} & \textbf{\color{red!70!black}134} & 19  \\
\rowcolor{blue!5}
Qwen2-VL-Instruct~\cite{Qwen2-VL} & 7B  & \textbf{\color{red!70!black}129} & 26 & 28 & 0 & \underline{\color{blue!70!black}0} & 27 \\
\rowcolor{blue!5}
DeepSeek-VL2~\cite{deepseek_vl2} & 16B & 30 & \textbf{\color{red!70!black}152} & \underline{\color{blue!70!black}0} & 2 & 0 & \textbf{\color{red!70!black}30}  \\
\rowcolor{blue!5}
Gemma 3~\cite{gemmateam2025gemma3technicalreport} & 27B  & 34 & 44 & \textbf{\color{red!70!black}110} & \underline{\color{blue!70!black}0} & 0 & 25  \\
\rowcolor{blue!5}
Mistral-3-Instruct~\cite{liu2026ministral} & 8B  & 109 & 64 & 13 & 2 & 0 & 26  \\
\rowcolor{blue!5}
UGround-V1~\cite{gou2024uground} & 7B & 62 & 66 & 55 & 4 & 0 & 25\\
\rowcolor{blue!5}
Seeclick~\cite{cheng2024seeclick} & - & 103 & 41 & 21 & \textbf{\color{red!70!black}11} & 2 & 29\\
\rowcolor{blue!5}
Qwen2.5-VL-Instruct~\cite{qwen2.5-VL} & 7B  & 20 & 40 & 78 & 10 & 10 & 26  \\
\rowcolor{blue!5}
CogAgent~\cite{hong2024cogagent} & 9B  & 57 & 43 & 36 & 3 & 16 & 26\\
\rowcolor{blue!5}
Aguvis~\cite{xu2025aguvis} & 7B & 42 & 84 & 31 & 4 & \underline{\color{blue!70!black}0} & 23 \\
\rowcolor{blue!5}
Os-Atlas~\cite{wu2024atlas} & 7B & 48 & 49 & 28 & 5 & 6 & 21\\
\rowcolor{blue!5}
UI-Tars 1.5-VL~\cite{qin2025ui}& 7B  & 20 & 55 & 30 & \underline{\color{blue!70!black}0} & \underline{\color{blue!70!black}0} & 22  \\
\rowcolor{blue!5}
Qwen3-VL-Instruct~\cite{qwen3vltechnicalreport} & 8B & \underline{\color{blue!70!black}16} & 35 & 40 & 6 & 21 & 18  \\
\rowcolor{blue!5}
GUI-Actor~\cite{wu2025gui} & 7B  & 26 & 26 & 21 & 1 & 24 & \underline{\color{blue!70!black}17}  \\

\midrule

\rowcolor{red!10}
\multicolumn{8}{l}{\textbf{Closed Source Models}} \\

\rowcolor{red!5}
GPT 4o mini~\cite{openai2024gpt4omini} & 8B--10B & 62 & 36 & 88 & 2 & 0 & 25\\
\rowcolor{red!5}
GPT 5~\cite{singh2025openai} & - & \underline{\color{blue!70!black}13} & 67 & 97 & 3 & 0 & 27 \\

\bottomrule
\end{tabular}
}

\label{tab:failure}
\end{table*}
\medskip
\noindent\textbf{6. Failure Taxonomy}
Failure analysis (Table~\ref{tab:failure}) reveals distinct structural weaknesses across model families. General-purpose \texttt{MLLMs} exhibit highly concentrated failure modes: LLaMA is dominated by \textit{No Prediction} errors (63\%), DeepSeek-VL2 by \textit{Far Miss} (71\%), and Qwen2-VL and Mistral-3 by \textit{Edge Bias} (64\% and 51\%), indicating systematic spatial misalignment in clinical interfaces. Closed-source models show similar patterns, with \texttt{GPT-5} heavily affected by \textit{Toolbar Confusion} (97) and \textit{Far Miss} (67), and GPT-4o-mini exhibiting substantial \textit{Toolbar Confusion} (88) and \textit{Edge Bias} (62). In contrast, interface-specialized models display more distributed, difficulty-driven errors: GUI-Actor spreads failures across \textit{Edge Bias}, \textit{Far Miss}, and \textit{Small Target}, while UI-TARS, Aguvis, and OS-Atlas show balanced spatial and semantic confusion patterns. Notably, \textit{Small Target} failures persist (12--18\%) across all models, reflecting the intrinsic challenge of compact toolbar icons in DICOM viewers. 

\begin{figure}[t]
    \centering
   \begin{subfigure}[t]{0.3\linewidth}
        \centering
        \includegraphics[width=\linewidth]{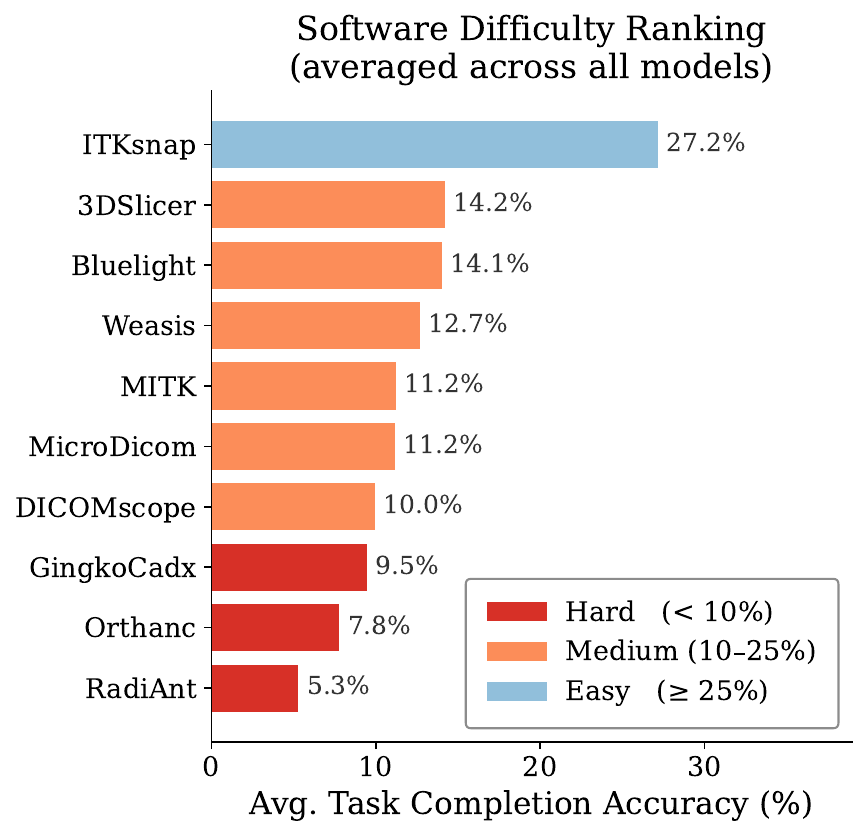}
        \caption{Software Difficulty Ranking}
        \label{fig:difficulty}
    \end{subfigure}
    \hfill
    \begin{subfigure}[t]{0.65\linewidth}
        \centering
        \includegraphics[width=\linewidth]{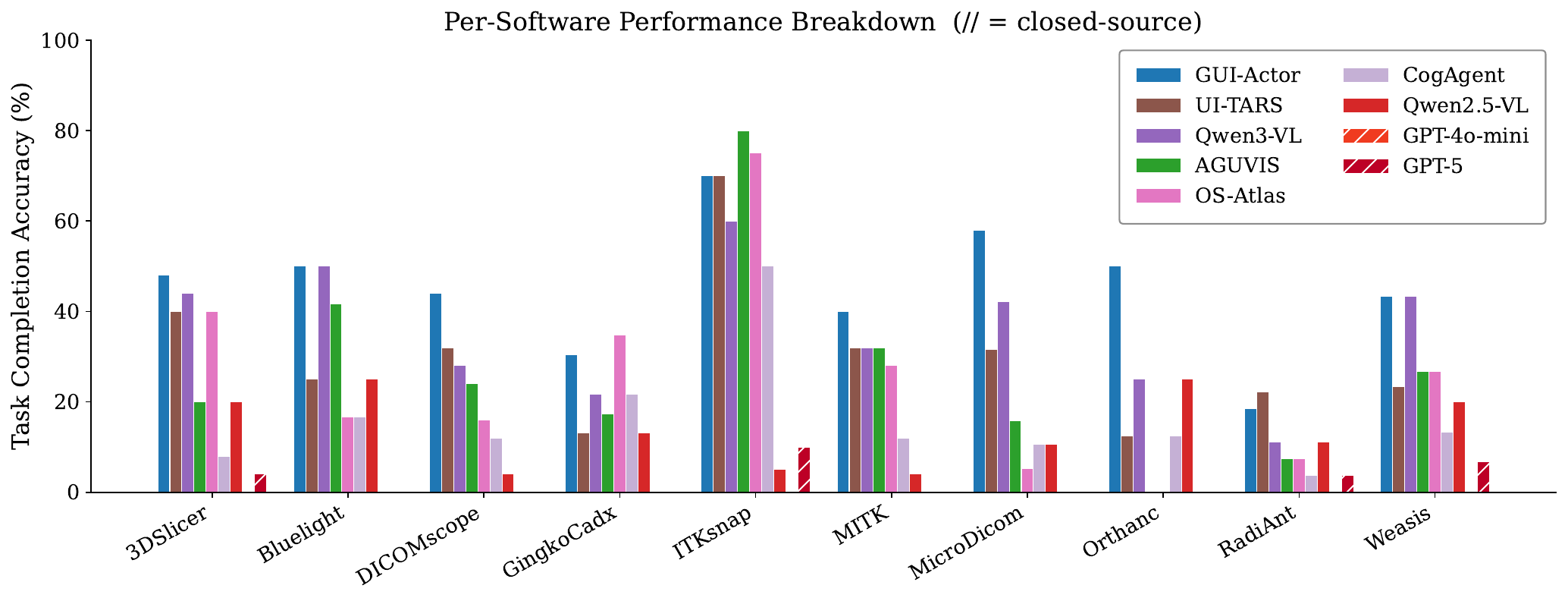}
        \caption{Per software performance}
        \label{fig:persoftware}
    \end{subfigure}
    \caption{
Per-software performance analysis. (a) Average \texttt{TCA} across models, ranking clinical applications by difficulty. (b) Per-model \texttt{TCA} breakdown across ten medical software platforms, highlighting interface-dependent performance variability.
}
    \label{fig:persoftware_combined}
\end{figure}

\subsection{Per-Software Performance Analysis}

\noindent\textbf{1. Software Difficulty Ranking.}
As shown in Fig.~\ref{fig:difficulty}, average Task Completion Accuracy (\texttt{TCA}) varies substantially across clinical applications, revealing strong interface-dependent performance sensitivity. RadiAnt (5.8\%) and Orthanc (below 10\%) emerge as the most challenging platforms, likely due to dense toolbar layouts and compact icon-heavy interfaces characteristic of DICOM viewers. Most applications fall within a moderate difficulty range (10--25\%), including DICOMscope, GinkgoCADx, MITK, MicroDicom, Weasis, 3D Slicer, and BlueLight. ITKsnap stands out as the least challenging application (30.4\% average \texttt{TCA}), benefiting from a comparatively structured and less cluttered interface layout. These findings confirm that layout density and interaction hierarchy significantly influence grounding reliability.

\noindent\textbf{2. Per-Software Breakdown Across Models.}
Figure~\ref{fig:persoftware} presents the per-model \texttt{TCA} breakdown across all ten applications. GUI-Actor consistently achieves the highest or near-highest performance across most platforms, peaking at 70\% on ITKsnap and maintaining strong results on MicroDicom (58\%) and Orthanc (50\%). UI-TARS follows a similar trend, also reaching 70\% on ITKsnap. AGUVIS achieves the single highest score (80\%) on ITKsnap but exhibits greater performance variance on denser interfaces such as RadiAnt and GinkgoCADx. While Qwen2.5-VL demonstrates moderate performance on 3D Slicer (20\%) and BlueLight (25\%), it collapses on DICOM-intensive tools, underscoring limited cross-interface robustness. Overall, no model exhibits uniform generalization across all software categories, reinforcing the importance of benchmark diversity and workflow-aware evaluation.
\textcolor{gray}{Insights are mentioned in the Supplementary material~\ref{sec:insights}.}

\section{Ablation Study}

We conduct systematic ablation experiments to isolate the impact of evaluation design and spatial stratification on \textbf{MedSPOT} performance. In each study, a single factor is varied while all other conditions remain fixed, enabling controlled analysis of causal effects.

\begin{figure}[t]
    \centering
    \begin{subfigure}[t]{0.49\linewidth}
        \centering
        \includegraphics[width=1\linewidth]{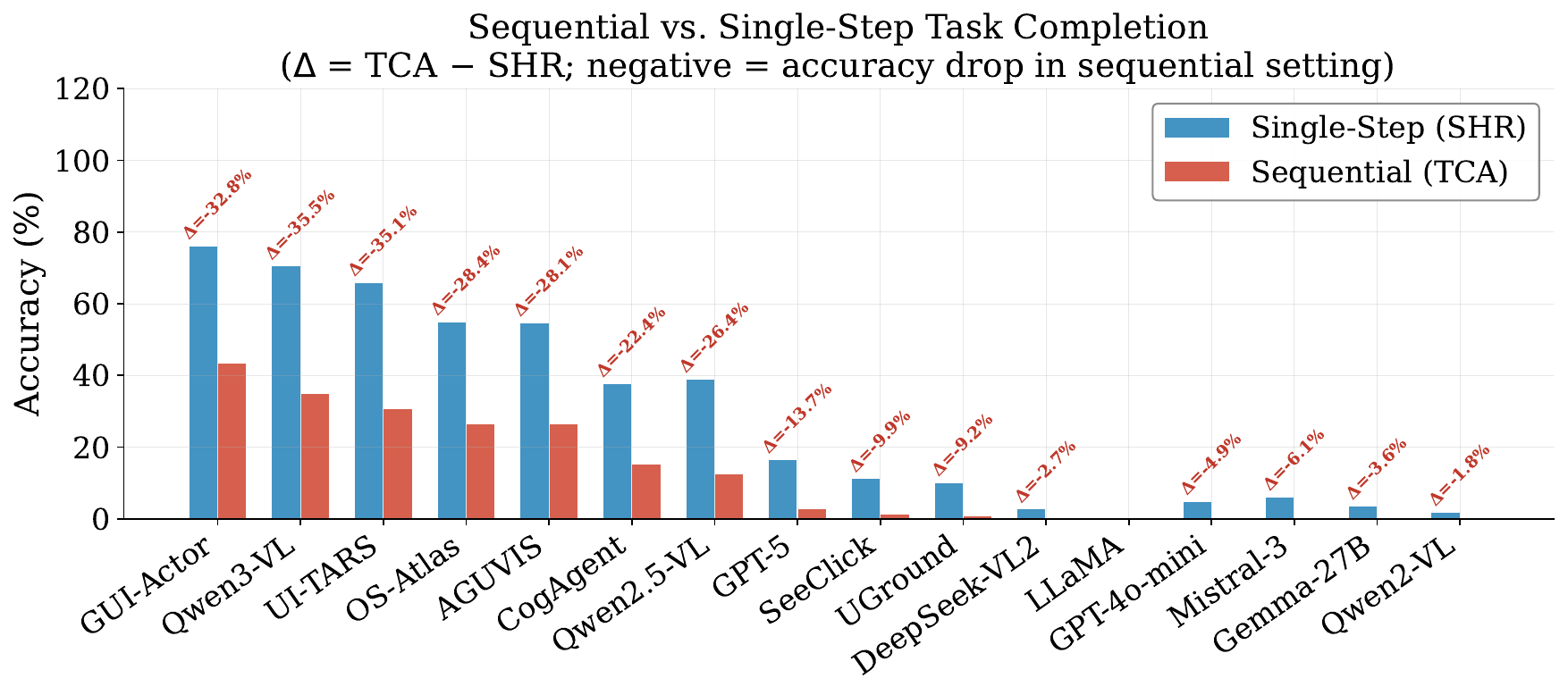}
        \caption{Sequential (\texttt{TCA}) vs. Single-Step (\texttt{SHR}) evaluation.}
        \label{fig:sse}
    \end{subfigure}
    \hfill
    \begin{subfigure}[t]{0.49\linewidth}
        \centering
        \includegraphics[width=1\linewidth]{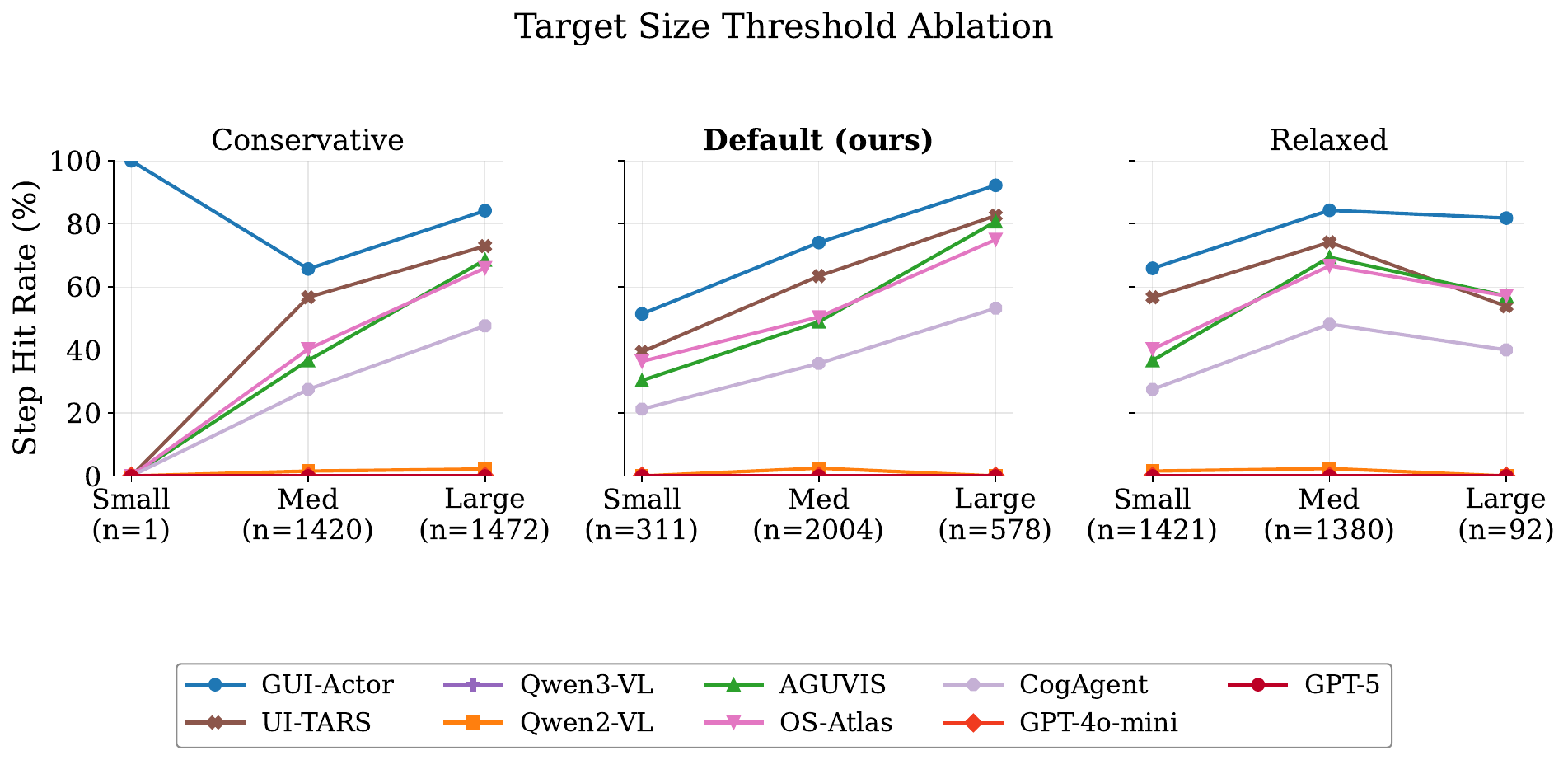}
        \caption{Step hit rate across size tiers under varying thresholds.}
        \label{fig:size}
    \end{subfigure}
    \caption{Ablation studies on evaluation design and spatial stratification. (a) Sequential evaluation significantly lowers task-level scores due to error compounding. (b) Small \texttt{UI} elements consistently exhibit lower accuracy across all threshold configurations.}
    \label{fig:ablation}
    \vspace{-0.5 cm}
\end{figure}

\subsection{Sequential vs. Single-Step Evaluation}

We compare two evaluation protocols: (i) \textbf{Sequential}, where a task is considered complete only if all steps succeed in order under early termination, and (ii) \textbf{Single-Step}, where each step is scored independently without enforcing temporal dependency.
As illustrated in Fig.~\ref{fig:sse}, Sequential evaluation (\texttt{TCA}) yields consistently lower scores than Single-Step Accuracy (\texttt{SHR}) across all models. The gap between \texttt{SHR} and \texttt{TCA} reflects compounding error under multi-step execution. Although some models show moderate single-step accuracy, their performance drops sharply under early termination, indicating that independent step scoring overestimates true task-level reliability.

\subsection{Impact of Target Size Thresholds}
Medical \texttt{GUI} elements range from large visualization panels to compact toolbar icons, introducing substantial variation in spatial difficulty. We stratify steps into \textit{small} ($<4{\times}10^{-4}$), \textit{medium} ($4{\times}10^{-4}$--$3{\times}10^{-3}$), and \textit{large} ($\ge3{\times}10^{-3}$) targets based on normalized bounding box area, and ablate these thresholds under \textit{Conservative}, \textit{Default}, and \textit{Relaxed} configurations (Fig.~\ref{fig:size}). While the threshold choice alters category distribution (from 1 to 1,421 small targets), performance ordering remains stable across settings: small elements are consistently the hardest to localize. All models exhibit marked degradation on compact UI components prevalent in dense \texttt{DICOM} viewers and annotation tools, confirming that the difficulty is intrinsic rather than threshold-induced. Notably, Qwen2-VL approaches near-zero accuracy across size regimes, underscoring insufficient spatial precision for fine-grained medical \texttt{GUI} grounding.

\subsection{Qualitative Analysis}

\begin{figure}[t]
    \centering
    \includegraphics[width=\linewidth]{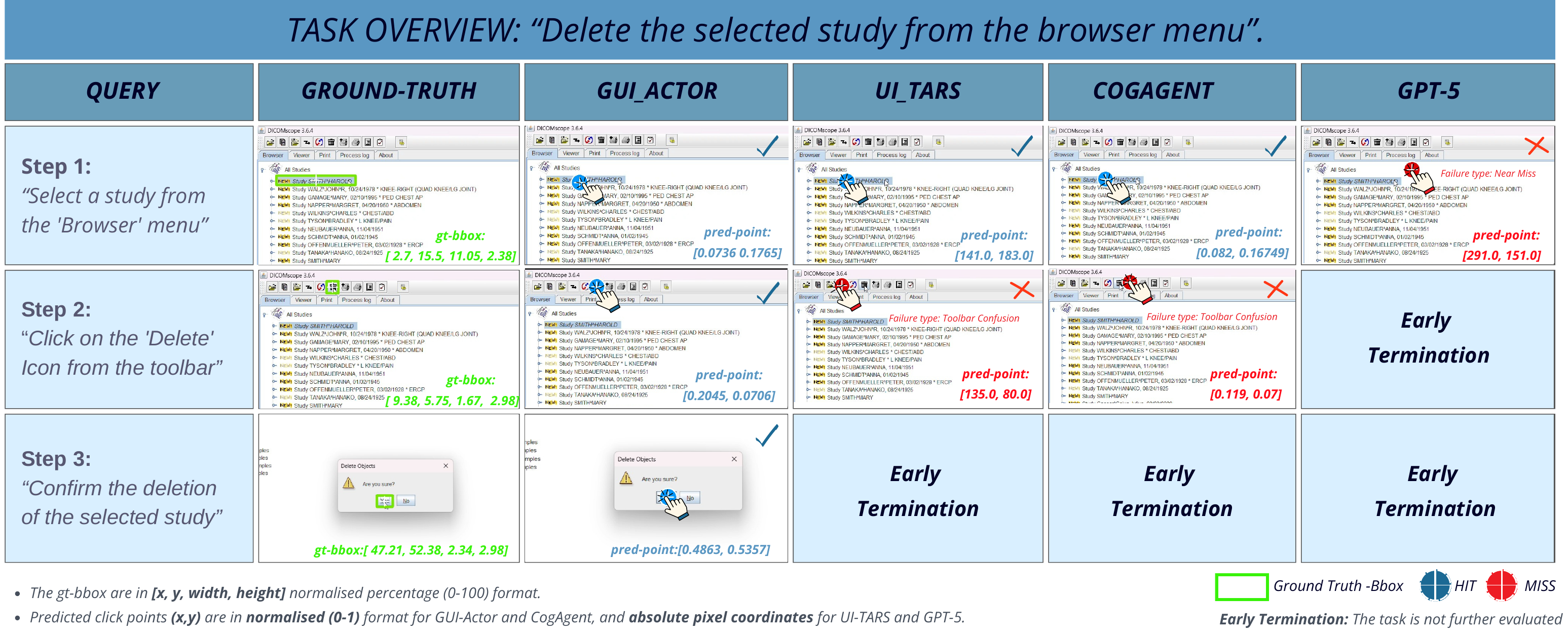}
    \caption{Qualitative comparison on the task \textit{"Delete the selected study from the browser menu"} (DICOMscope). GUI-Actor completes all steps, while other models fail due to toolbar confusion or near-miss predictions, resulting in early termination. Green box = ground truth; \textcolor{blue}{blue} = correct prediction; \textcolor{red}{red} = incorrect prediction. \textcolor{gray}{Further examples are in the Supplementary material~\ref{sec:qualitative_appendix}.}}
    \label{fig:qualitative}
    \vspace{- 0.3 cm}
\end{figure}

To illustrate sequential fragility in practice, Fig.~\ref{fig:qualitative} presents a representative DICOMscope task. GUI-Actor successfully completes all three interaction steps, while UI-TARS and CogAgent fail at step 2 due to toolbar confusion, triggering early termination. GPT-5 fails at the first step with a near-miss prediction and cannot recover. These examples highlight a key insight: partial grounding competence does not imply workflow-level robustness under state dependency.

\section{Conclusion}

We introduced \textbf{MedSPOT}, the first benchmark for workflow-aware spatial grounding in clinical software interfaces. Spanning 216 task-driven videos and 597 annotated keyframes across 10 medical platforms, \textbf{MedSPOT} models grounding as a sequential, state-dependent process rather than isolated click prediction. 
Our strict early-termination protocol reveals substantial fragility in current multimodal models: even the strongest architecture achieves only 43.5\% Task Completion Accuracy, highlighting the intrinsic difficulty of sustained grounding in dense clinical \texttt{GUIs}. Through comprehensive evaluation and failure taxonomy analysis, we demonstrate that performance degradation arises from systematic spatial and interface-structure challenges rather than incidental errors.
\textbf{MedSPOT} establishes a realistic and safety-critical benchmark for multimodal interaction research in healthcare software, and we release all data and evaluation tools to support future advances in workflow-aware medical automation.

\noindent\textbf{Limitations.}
\textbf{MedSPOT} is currently monolingual, limited in scale (216 video tasks), and supports only click-based interactions, excluding common actions such as drag, scroll, and text input. Evaluation is restricted to spatial grounding by checking whether predicted coordinates fall within ground-truth \texttt{GUI} bounding boxes, abstracting richer interaction semantics. Additionally, the strict early-termination protocol, while realistic, may underestimate partial procedural competence by invalidating subsequent steps after a single error. \textcolor{gray}{Detailed Limitations in Supplementary Material~\ref{sec:limitations_appendix}.}

%
%
\bibliographystyle{splncs04}
\bibliography{main}

\appendix
\renewcommand{\thesection}{\Alph{section}}
\setcounter{section}{0}

\newpage
\chapter*{Supplementary Material: 
MedSPOT: A Workflow-Aware Sequential
Grounding Benchmark for Clinical GUI (Paper 14458)}

\begin{table}[ht]
\centering
\renewcommand{\arraystretch}{1.35}

\resizebox{\linewidth}{!}{

\rowcolors{2}{gray!8}{white}

\begin{tabular}{@{}p{4cm} p{9.5cm}@{}}

\toprule
\rowcolor{blue!20}
\textbf{Appendix Section} & \textbf{Description} \\
\midrule

\texttt{Section~\ref{datasetcard}} & \textbf{MedSPOT Dataset Card} \\

\texttt{Section~\ref{sec:inferencep}} & \textbf{Inference Protocol} \\

\texttt{Section~\ref{sec:stats}} & \textbf{Statistical Robustness} \\

\texttt{Section~\ref{sec:ethics}} & \textbf{Data Provenance \& Ethics} \\

\texttt{Section~\ref{sec:failure}} & \textbf{Failure Taxonomy Definitions} \\

\texttt{Section~\ref{sec:insights}} & \textbf{Insights} \\

\texttt{Section~\ref{sec:limitations_appendix}} & \textbf{Limitations: Visual Examples} \\

\texttt{Section~\ref{sec:qualitative_appendix}} & \textbf{Qualitative Analysis: Visual Examples} \\

\texttt{Section~\ref{sec:iaa}} & \textbf{Inter-Annotator Agreement} \\

\bottomrule

\end{tabular}
}

\label{tab:appendix_contents}

\end{table}

\section{MedSPOT DataCard}
\phantomsection
\label{datasetcard}

\subsection{Motivation}

\begin{itemize}

\item \textbf{Purpose of the dataset.}

\textbf{MedSPOT} was created to address the lack of benchmarks for \textit{workflow-aware sequential GUI grounding} in clinical software environments. Existing grounding benchmarks typically evaluate single-step interactions on general-purpose applications. In contrast, clinical workflows require models to perform a sequence of causally dependent spatial decisions across evolving interface states.  
\textbf{MedSPOT} fills this gap by modeling grounding as a multi-step reasoning process within real clinical GUI systems.

\item \textbf{Dataset creators.}

The dataset was created by the authors of this work as part of an academic research effort on evaluating multimodal grounding capabilities in clinical software interfaces.

\end{itemize}

\subsection{Dataset Composition}

\begin{table}[h!]
\centering
\caption{\textbf{MedSPOT} dataset composition.}
\label{tab:dataset_composition}

\renewcommand{\arraystretch}{1.3}

\rowcolors{2}{tablerow}{white}

\begin{tabular}{p{5cm} p{6cm}}

\toprule
\rowcolor{headerblue}
\textbf{Property} & \textbf{Value} \\
\midrule

Tasks & 216 task-driven interaction sequences \\
Annotated Keyframes & 597 \\
Software Platforms & 10 \\
Average Steps per Task & 2--3 \\
Imaging Modalities & CT, MRI, PET, X-Ray, Ultrasound \\

\midrule
\rowcolor{headerblue}
\multicolumn{2}{c}{\textbf{Interface Categories}} \\
\midrule

DICOM / PACS Viewers & 109 tasks \\
Segmentation \& Research Tools & 70 tasks \\
Web-Based Viewers & 37 tasks \\

\bottomrule
\end{tabular}

\end{table}

Table~\ref{tab:dataset_composition} summarizes the dataset composition.  
Unlike traditional grounding benchmarks that evaluate isolated actions, each task in \textbf{MedSPOT} consists of 2–3 sequential steps that must be executed correctly across evolving GUI states.

\subsection{Software Platforms}

Table~\ref{tab:software_benchmark} summarizes the ten clinical imaging software platforms included in \textbf{MedSPOT}. These platforms span a diverse set of application categories, including DICOM servers, image viewers, segmentation tools, and web-based medical imaging interfaces. Collectively, they support multiple medical imaging modalities such as CT, MRI, PET, ultrasound, and X-ray.  
By incorporating software with different interface layouts, interaction paradigms, and licensing models, \textbf{MedSPOT} captures a wide range of realistic clinical GUI environments. This diversity helps ensure that the benchmark evaluates grounding performance across heterogeneous medical imaging systems rather than being tailored to a single interface design.

\begin{table}[t]
\centering
\small
\setlength{\tabcolsep}{6pt}
\renewcommand{\arraystretch}{1.3}
\caption{Dataset curation and benchmark software platforms included in \textbf{MedSPOT}.}
\rowcolors{2}{rowgray}{white}

\resizebox{0.9\textwidth}{!}{
\begin{tabular}{@{}c c c c c@{}}

\toprule
\rowcolor{headerblue}
\textbf{Software} & \textbf{Type} & \textbf{Tasks} & \textbf{Modalities} & \textbf{License} \\
\midrule

Orthanc~\cite{Jodogne2018} & Server & 10 & CT, MRI, Ultrasound, X-ray, PET & GPL \\

ITK-SNAP~\cite{py06nimg} & Segmentation & 20 & CT, MRI, CBCT, NIfTI & GPL \\

Weasis~\cite{weasis} & Viewer & 30 & CT, MRI, Ultrasound, X-ray, PET & EPL 2.0 \\

Bluelight~\cite{chen2023bluelight} & Web Viewer & 12 & CT, MRI, Ultrasound, X-ray & MIT \\

MITK~\cite{wolf2005} & Viewer / Seg. & 25 & CT, MRI, Ultrasound, PET & BSD \\

RadiAnt~\cite{radiant} & Viewer & 27 & CT, MRI, US, X-ray, PET & Freeware \\

MicroDICOM~\cite{MicroDicom} & Viewer & 19 & CT, MRI, Ultrasound, X-ray & Freeware \\

Ginkgo CADx~\cite{ginkgocadx} & Viewer / Analysis & 23 & CT, MRI, PET, Ultrasound, X-ray & LGPL \\

DICOMscope~\cite{dicomscope} & Viewer & 25 & CT, MRI, Ultrasound, Angio & BSD \\

3D Slicer~\cite{slicer2015} & Viewer / Seg. & 25 & CT, MRI, Ultrasound, 3D volumes & BSD \\

\bottomrule
\end{tabular}
}

\label{tab:software_benchmark}

\end{table}

\subsection{Annotation Protocol}

The annotation process in \textbf{MedSPOT} is designed to capture the sequential and state-dependent nature of interactions within clinical graphical user interfaces (GUIs). Each annotated keyframe corresponds to a distinct interaction step within a task-driven workflow. For every such step, we record a structured annotation tuple:

\begin{equation}
A_t = (I_t,\; s_t,\; y_t,\; B_t)
\end{equation}

where each component represents a different aspect of the grounding problem:

\begin{itemize}

\item $I_t$ denotes the GUI screenshot captured at interaction step $t$. This image represents the full interface state at the moment when the target action must be performed. Screenshots preserve the spatial layout of interface elements such as menus, toolbars, buttons, and dialogs.

\item $s_t$ is the natural-language instruction associated with the interaction step. This instruction describes the action that should be performed in the interface (e.g., \textit{“Click the Open File icon”} or \textit{“Select the Fire LUT from the palette”}). These instructions emulate the form of commands that might be given to a multimodal model performing GUI-based tasks.

\item $y_t$ represents the semantic description of the target element within the interface. This field provides a concise textual identifier for the interface component to be grounded (e.g., \textit{Open File icon}, \textit{Help menu}, \textit{Apply button}). The semantic description serves as an additional grounding signal linking the instruction to the visual element.

\item $B_t = (x, y, w, h)$ denotes the ground-truth bounding box of the target GUI element within the screenshot. The bounding box specifies the spatial location of the actionable element that satisfies the instruction. Bounding boxes are expressed using normalized percentage coordinates in the range $[0,100]$, where $(x,y)$ indicates the top-left corner of the box and $(w,h)$ denote its width and height relative to the image dimensions.

\end{itemize}

Using normalized coordinates ensures that annotations remain consistent across screenshots with different resolutions and display configurations. This representation allows models to focus on relative spatial reasoning within the interface rather than absolute pixel positions.

The annotation protocol follows the temporal structure of real interaction workflows. Each task is decomposed into a sequence of GUI states, where every step depends on the successful completion of the previous action. Consequently, the dataset captures not only individual grounding instances but also the causal dependencies between successive interface states.

Figures~\ref{fig:taskexample0}–\ref{fig:taskexample3} illustrate representative examples of annotated tasks and the corresponding GUI states extracted from interaction videos. These examples demonstrate how instructions, semantic targets, and bounding box annotations align across sequential interface transitions.

\newpage
\subsection{Data Examples}

\begin{figure}[htbp]
\centering
\includegraphics[width=\linewidth]{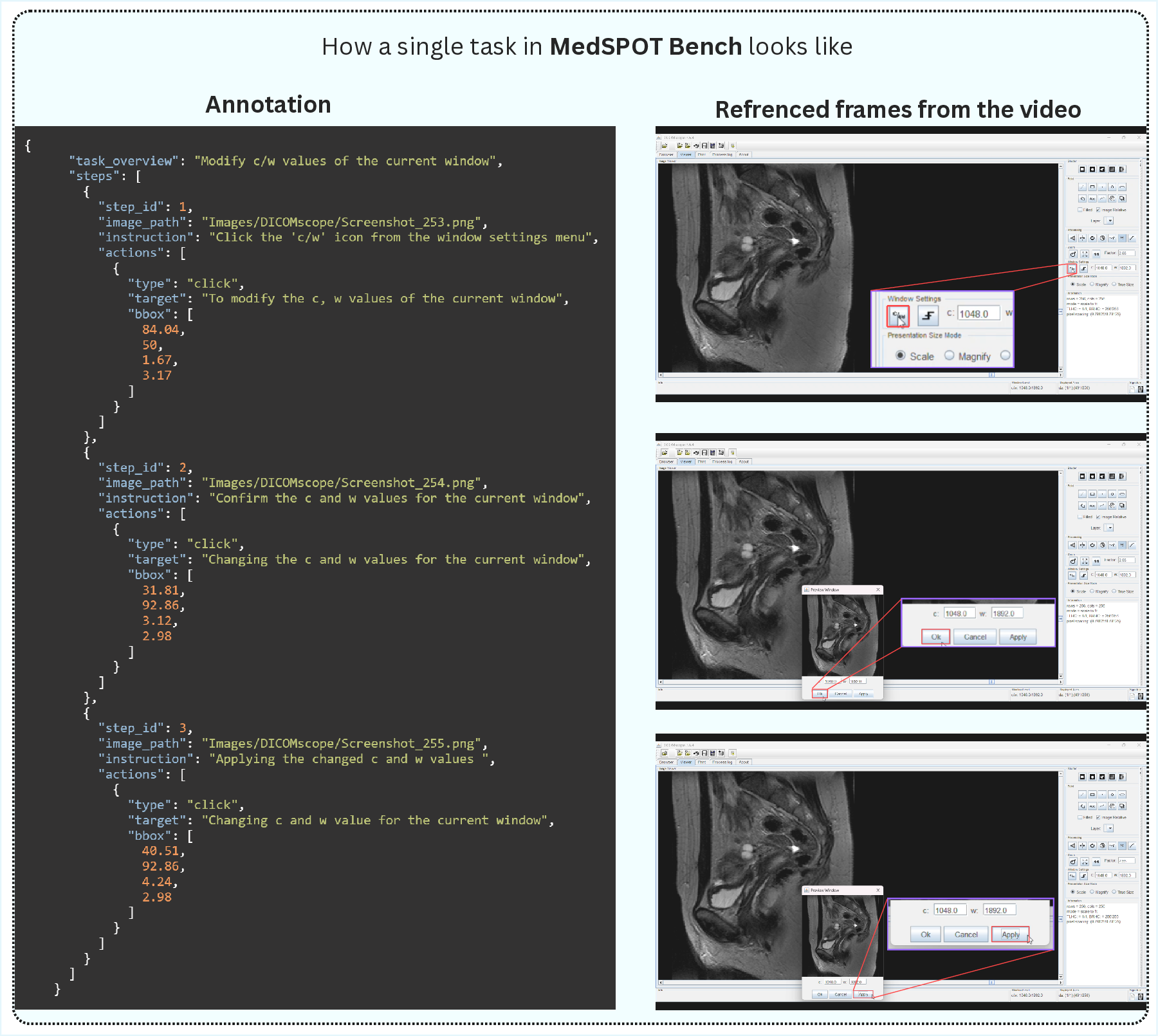}
\caption{\textbf{Window-level adjustment workflow.} This example illustrates a three-step interaction for modifying image windowing parameters. (1) The model must first locate and click the \texttt{c/w} icon in the viewer toolbar to open the window-level adjustment dialog. (2) Once the dialog appears, the model must confirm the selected center and width parameters by clicking the \texttt{OK} button. (3) Finally, the model applies the changes by clicking the \texttt{Apply} button. This example demonstrates a multi-stage GUI workflow in which each step reveals a new interface state.}
\label{fig:taskexample0}
\end{figure}

\begin{figure}[htbp]
\centering
\includegraphics[width=\linewidth]{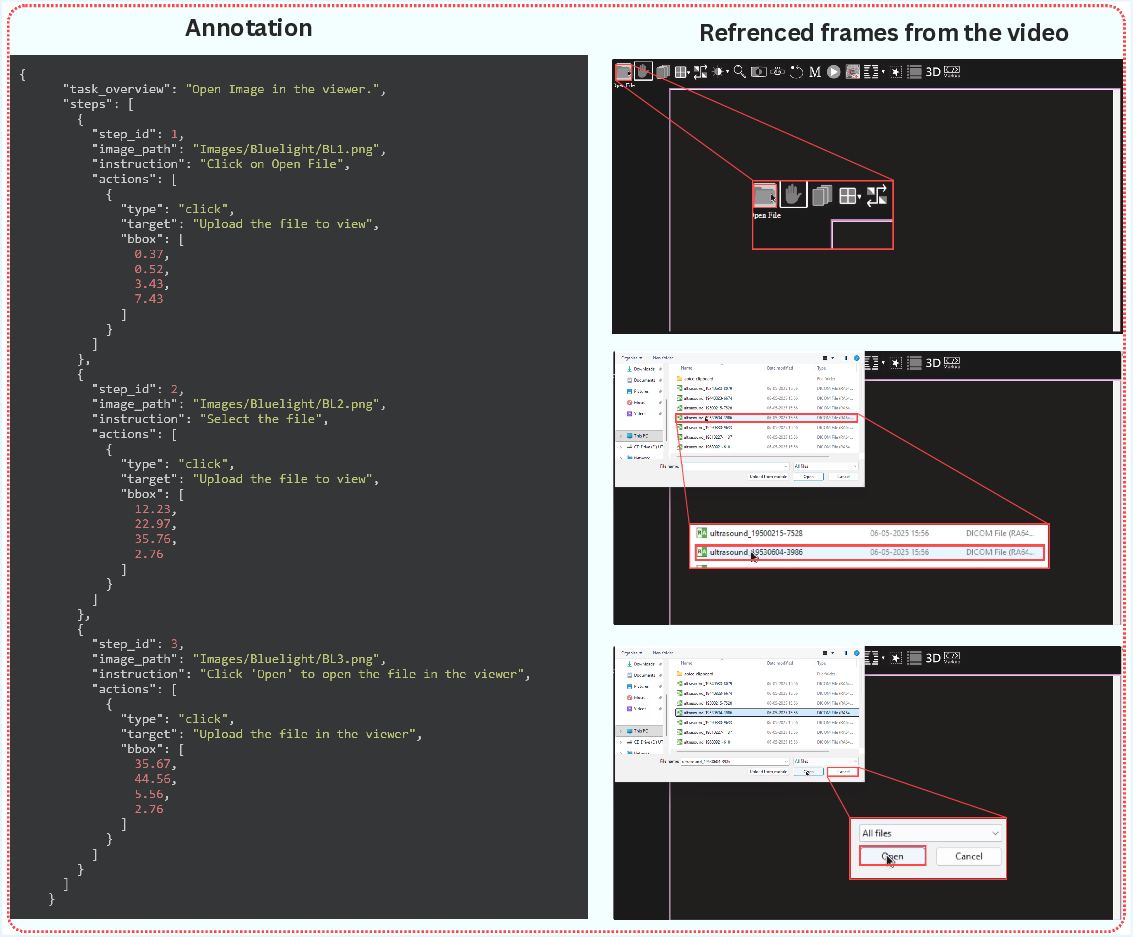}
\caption{\textbf{File loading workflow.} The task proceeds through three sequential actions. (1) The model must first click the \texttt{Open File} icon located in the viewer toolbar, which launches the system file explorer. (2) Within the file explorer interface, the model must identify and select the desired DICOM file from the list of available files. (3) Finally, the model confirms the selection by clicking the \texttt{Open} button to load the image into the viewer. This example highlights cross-interface grounding between application windows.}
\label{fig:taskexample1}
\end{figure}

\begin{figure}[htbp]
\centering
\includegraphics[width=\linewidth]{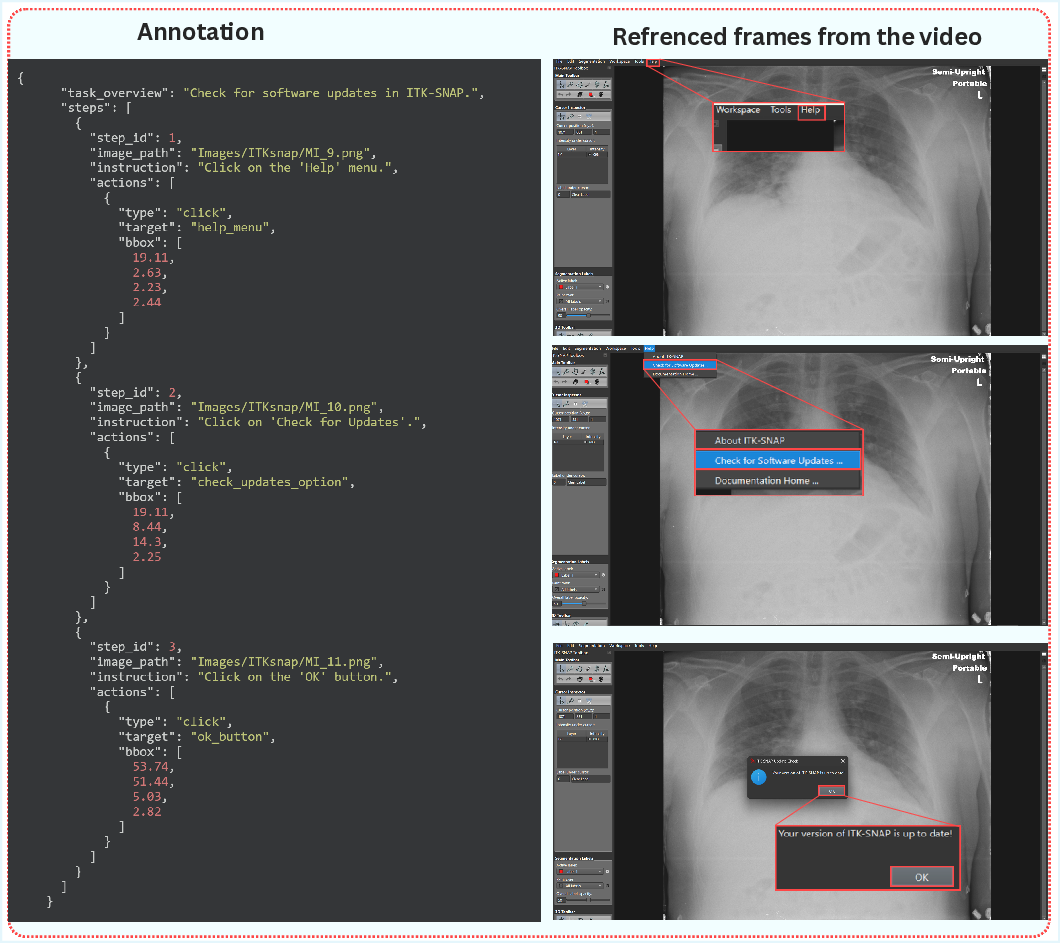}
\caption{\textbf{Software update notification workflow.} This task involves navigating a menu hierarchy. (1) The model first identifies and clicks the \texttt{Help} menu in the top menu bar. (2) From the dropdown options that appear, it selects \texttt{Check for Software Updates}. (3) After the update status dialog appears, the model confirms the notification by clicking the \texttt{OK} button. This example illustrates hierarchical menu navigation within a GUI.}
\label{fig:taskexample2}
\end{figure}

\begin{figure}[htbp]
\centering
\includegraphics[width=\linewidth]{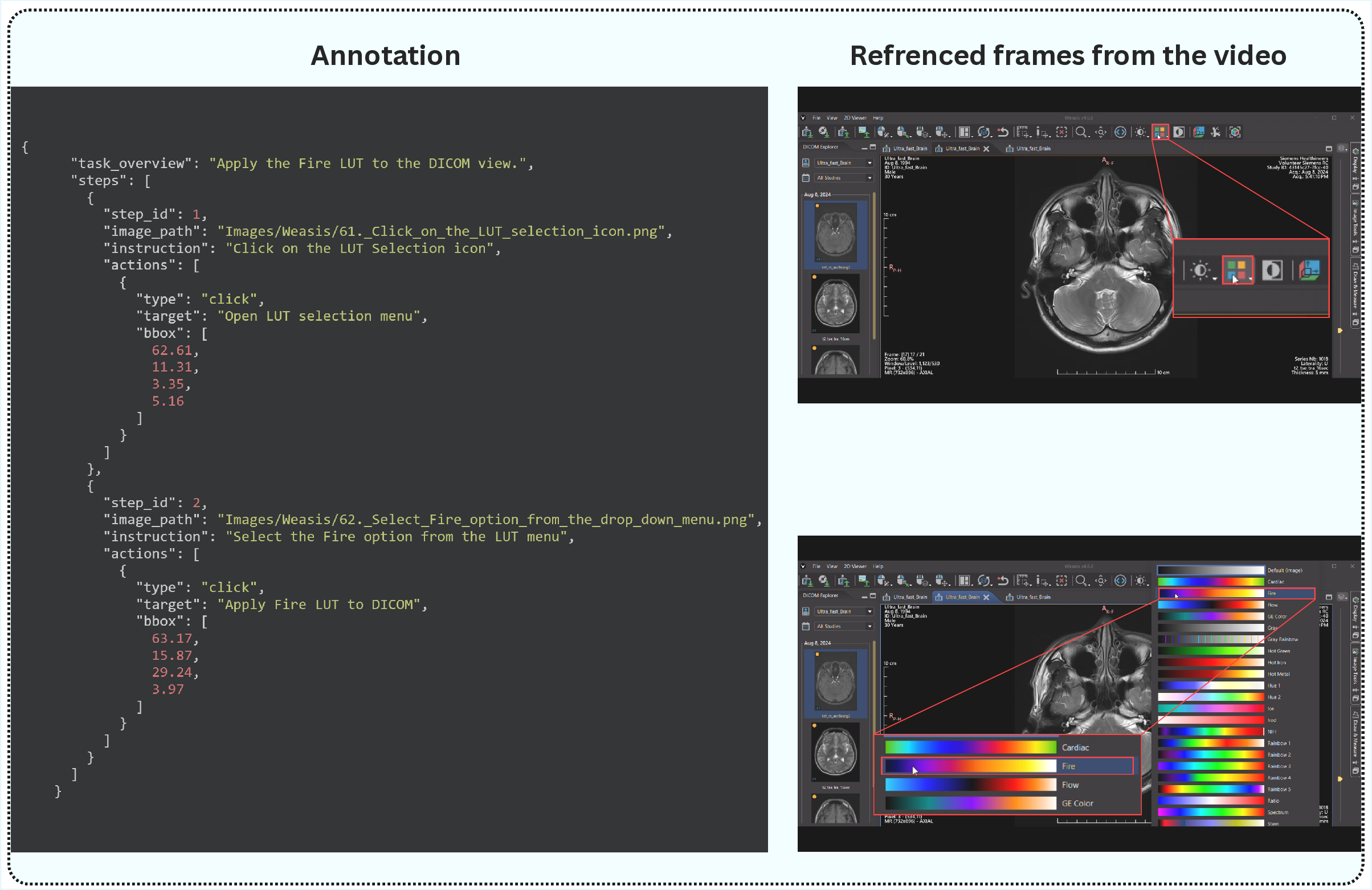}
\caption{\textbf{Lookup table (LUT) selection workflow.} This task demonstrates modification of visualization settings in a medical imaging viewer. (1) The model first opens the LUT selection menu by clicking the LUT icon in the viewer toolbar. (2) From the available color mapping options in the dropdown palette, the model selects the \texttt{Fire} lookup table, which updates the rendering of the medical image.}
\label{fig:taskexample3}
\end{figure}

\begin{figure}[htbp]
\centering
\includegraphics[width=0.8\linewidth]{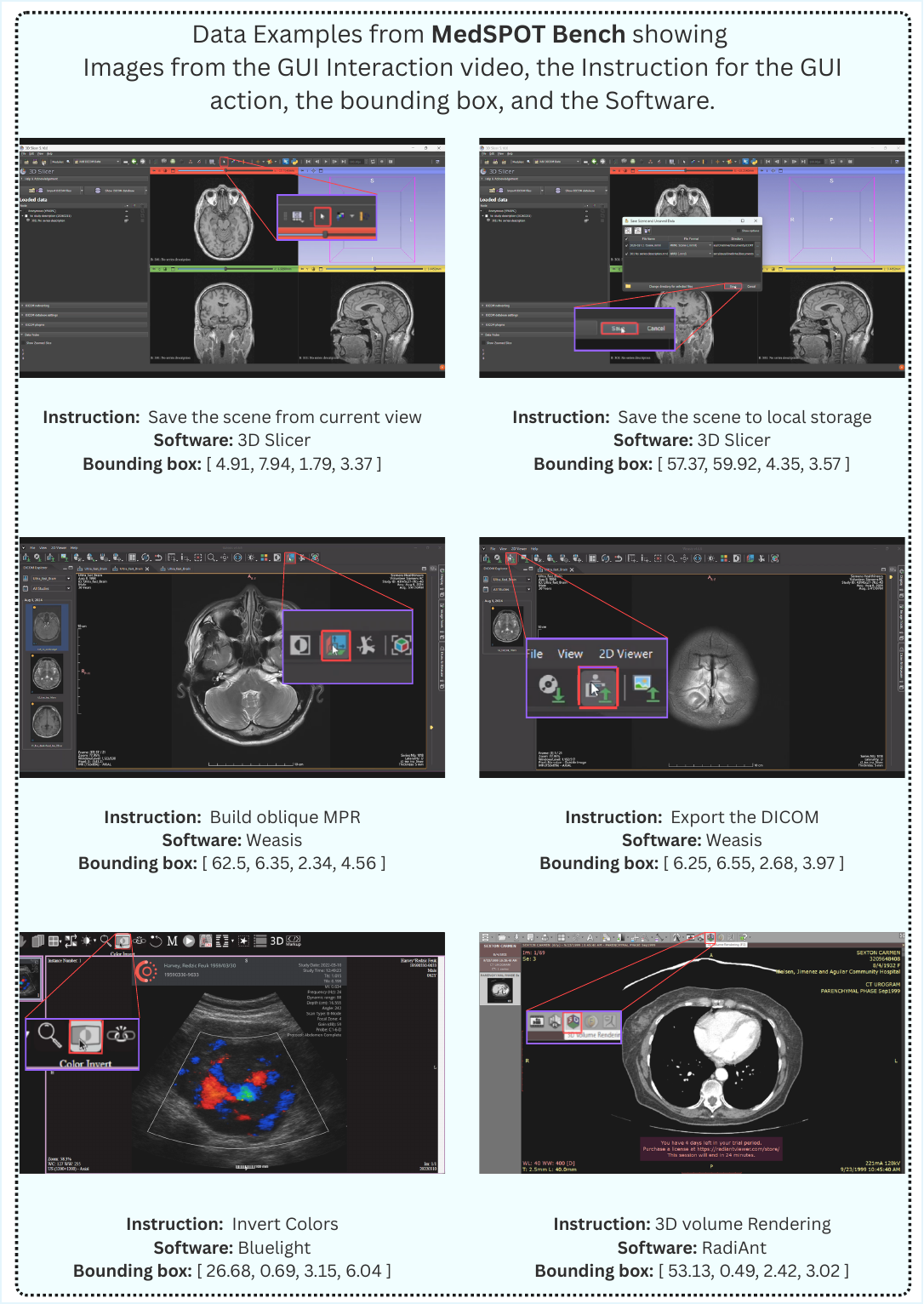}
\caption{\textbf{Example annotation structure in MedSPOT.} Each annotated step contains the GUI screenshot, the associated natural language instruction, a semantic description of the target GUI element, and the ground-truth bounding box identifying the element to be interacted with. Bounding boxes are represented using normalized coordinates, enabling consistent annotation across screenshots with different resolutions.}
\label{fig:dataexamples}
\end{figure}

To illustrate the structure of tasks and annotations in \textbf{MedSPOT}, Figures~\ref{fig:taskexample0}--\ref{fig:taskexample3} present representative examples of sequential GUI interactions extracted from the dataset. 
Figure~\ref{fig:dataexamples} further illustrates the annotation structure used in \textbf{MedSPOT}. For each step, the dataset provides the GUI screenshot, the natural language instruction describing the intended action, the semantic description of the target interface element, and the ground-truth bounding box indicating the spatial location of the element to be clicked. Together, these components form the structured annotation tuple described in Section~\ref{datasetcard}.

\newpage
\subsection{Recording and Labeling Process}

Interaction sessions were recorded at native screen resolution using standard screen-capture software to preserve the original layout and spatial structure of the graphical user interfaces (GUIs). From these recordings, keyframes were extracted at meaningful interface state transitions—specifically when a user action triggered a visible change in the interface (e.g., opening a dialog, expanding a menu, or loading a file). Each extracted frame therefore corresponds to a distinct step in a task-driven workflow.

Annotations were performed using \textbf{Label Studio}. A team of three annotators manually labeled the spatial location of the target GUI element for every step by drawing bounding boxes around the actionable interface component. To ensure spatial accuracy and consistency, each annotation was verified against the original interaction video. If a GUI state appeared ambiguous or the intended interaction target was unclear, the corresponding task sequence was excluded from the final dataset.
This process resulted in a curated collection of high-quality annotations that preserve both the spatial grounding signal and the sequential dependencies between interaction steps.

\subsection{Intended Use}

\textbf{MedSPOT} is designed as a research benchmark for evaluating the spatial grounding capabilities of multimodal large language models (MLLMs) in clinical GUI environments. The benchmark focuses on sequential GUI interactions, requiring models to correctly identify interface elements across evolving application states.
The dataset is intended strictly for research and evaluation purposes. Current model performance remains far below the reliability required for clinical deployment. For instance, the best-performing model in our evaluation achieves only \textbf{43.5\% task completion accuracy (TCA)}, highlighting the substantial challenges that remain in developing reliable GUI-grounded reasoning systems.

\section{Inference Protocol}
\phantomsection
\label{sec:inferencep}

\subsection{General Setup}

All open-source models, except \textbf{Mistral} and \textbf{DeepSeek}, were evaluated on a single NVIDIA A10 GPU with 24GB memory. Due to their larger memory requirements, \textbf{Mistral} and \textbf{DeepSeek} were evaluated on a single NVIDIA H100 GPU with 100GB memory. Closed-source models were accessed via their respective APIs.
For all experiments, input images were provided at their native screen resolution without any resizing or preprocessing. This design choice preserves the spatial layout of GUI elements and ensures that evaluation reflects realistic user-interface conditions.

\subsection{Per-Model Configuration}

\begin{table}[t]
\centering
\caption{Per-model inference configuration used for evaluation on \textbf{MedSPOT}.}
\label{tab:inference}

\small
\setlength{\tabcolsep}{5pt}
\renewcommand{\arraystretch}{1.25}

\rowcolors{2}{rowgray}{white}

\resizebox{\textwidth}{!}{
\begin{tabular}{l l c c c c}

\toprule
\rowcolor{tableheader}
\textbf{Model} & \textbf{Checkpoint} & \textbf{Precision} & \textbf{Coord. Format} & \textbf{K} & \textbf{Temp} \\
\midrule

GUI-Actor & microsoft/GUI-Actor-7B-Qwen2.5-VL & BF16 & Norm [0,1] & 5 & 0 \\
UI-TARS & bytedance/UI-TARS-7B-SFT & BF16 & Absolute px & 1 & 0.3 \\
OS-Atlas & OS-Copilot/OS-Atlas-Base-7B & BF16 & [0,1000] & 1 & 0 \\
CogAgent & CogAgent-9B & BF16 & [0,1000] & 1 & 0 \\
AGUVIS & AGUVIS-7B & BF16 & Norm [0,1] & 1 & 0 \\
SeeClick & SeeClick & BF16 & Norm [0,1] & 1 & 0 \\
UGround & UGround-V1-7B & BF16 & Norm [0,1] & 1 & 0 \\
Qwen2-VL & Qwen2-VL-7B-Instruct & BF16 & Absolute px & 1 & 0 \\
Qwen2.5-VL & Qwen2.5-VL-7B-Instruct & BF16 & Absolute px & 1 & 0 \\
Qwen3-VL & Qwen3-VL-8B-Instruct & BF16 & Absolute px & 1 & 0 \\
Llama 3.2 & Llama-3.2-11B-Vision-Instruct & BF16 & Absolute px & 1 & 0 \\
Gemma 3 & gemma-3-27b-it & BF16 & Absolute px & 1 & 0 \\
DeepSeek-VL2 & deepseek-vl2 & BF16 & Absolute px & 1 & 0 \\
Mistral-3 & Mistral-Small-3.1-24B-Instruct & BF16 & Absolute px & 1 & 0 \\
GPT-4o-mini & gpt-4o-mini & -- & Absolute px & 1 & 0 \\
GPT-5 & gpt-5 & -- & Absolute px & 1 & 0 \\

\bottomrule
\end{tabular}
}

\end{table}

Table ~\ref{tab:inference} shows the details of the model inference configuration for every model evaluated on the \textbf{MedSPOT Bench}.
GUI-Actor produces multiple candidate coordinates by design, as its architecture explicitly models spatial uncertainty over the target UI region rather than committing to a single deterministic prediction. Accordingly, we evaluate GUI-Actor using $K = 5$ candidate coordinates, consistent with its default inference configuration. This is analogous to evaluating beam search models using the full beam rather than restricting evaluation to the greedy top-1 output. In contrast, all other models generate a single coordinate by design and are therefore evaluated with $K = 1$.

During evaluation, a step for GUI-Actor is considered correct if any of the $K$ predicted coordinates falls within the ground truth bounding box $B_t$, or if a $\pm 14$px patch centered at any prediction overlaps with $B_t$. All other models are evaluated under the $K = 1$ setting. We acknowledge this difference in evaluation protocol. For completeness, we also report results under the stricter $K = 3$ setting used in the official ScreenSpot evaluation of GUI-Actor~\cite{wu2025gui}. Under this protocol, GUI-Actor achieves $\mathrm{TCA} = 33.64\%$, remaining the top-performing model. When restricted to $K = 1$, its $\mathrm{TCA}$ decreases to $28.04\%$. Importantly, the relative ranking among the remaining models remains unchanged under both evaluation settings.

\subsection{Prompt Templates}

\textbf{1. GUI-Specialized Models} 
(GUI-Actor, OS-Atlas, AGUVIS, SeeClick, UGround)

\begin{tcolorbox}[colback=promptblue,colframe=blue!70!black,title=Prompt Template]
You are a GUI agent. Locate the UI element described by the instruction and click inside it.

Instruction: \{instruction\}
\end{tcolorbox}

\textbf{2. General-Purpose Models} 
(Qwen2-VL, Qwen2.5-VL, Qwen3-VL, Llama 3.2, Gemma 3, DeepSeek-VL2, Mistral-3)

\begin{tcolorbox}[colback=promptblue,colframe=blue!70!black,title=Prompt Template]
You are a GUI grounding assistant. Given the screenshot, predict the click coordinate (x, y) of the UI element described.

Output only the coordinate.

Instruction: \{instruction\}
\end{tcolorbox}

\textbf{3. Closed-Source Models} 
(GPT-4o-mini, GPT-5)

\begin{tcolorbox}[colback=promptblue,colframe=blue!70!black,title=Prompt Template]
You are a GUI agent. Given a screenshot and an instruction, return the click coordinate as JSON:

\{``x'': value, ``y'': value\}

Use absolute pixel coordinates.

Instruction: \{instruction\}
\end{tcolorbox}

\textbf{4. CogAgent}

\begin{tcolorbox}[colback=promptblue,colframe=blue!70!black,title=Prompt Template]
I am using \{software\}.

\{instruction\}

Please provide the normalized coordinate.
\end{tcolorbox}
\subsection{Coordinate Extraction}

Different models generate predicted coordinates using heterogeneous output formats and spatial scales . To ensure fair and consistent evaluation across models, we implement model-specific parsing logic that converts all predicted coordinates into a unified normalized coordinate space.
Regardless of the original representation, every predicted coordinate is converted into normalized $[0,1]$ space before comparison with the ground-truth bounding box $B_t$. This normalization allows models operating in pixel space, fixed integer ranges, or normalized outputs to be evaluated under the same spatial metric.

For reasoning-oriented models such as \textbf{UI-TARS}, \textbf{Qwen2-VL}, \textbf{Qwen2.5-VL}, and \textbf{Qwen3-VL}, outputs frequently contain chain-of-thought reasoning traces in which intermediate coordinates may appear before the final answer. To avoid incorrectly extracting these intermediate values, we apply a \textit{think-block stripping} step prior to parsing. This step removes reasoning text and isolates the final predicted coordinate before regex-based extraction.
If coordinate extraction fails entirely for a given step, the prediction is recorded as \textbf{No Prediction} in the failure taxonomy.
Table~\ref{tab:coordinateextraction} summarizes the model-specific coordinate extraction and normalization strategies applied during evaluation.

\begin{table}[t]
\centering
\caption{Model-specific coordinate extraction and normalization procedures used in evaluation.}
\label{tab:coordinateextraction}

\small
\renewcommand{\arraystretch}{1.25}
\setlength{\tabcolsep}{6pt}

\rowcolors{2}{rowshade}{white}

\resizebox{\textwidth}{!}{
\begin{tabular}{c c c}

\toprule
\rowcolor{tableheader}
\textbf{Output Format} & \textbf{Models} & \textbf{Normalization Strategy} \\
\midrule

Normalized $[0,1]$ 
& AGUVIS, UGround 
& No normalization required \\

$[0,1000]$ range 
& OS-Atlas, CogAgent, SeeClick, Qwen3-VL 
& Divide coordinates by $1000$ \\

Absolute pixels 
& GPT-4o-mini, GPT-5, Gemma-3, LLaMA 
& Divide by image width $W$ and height $H$ \\

Native pixels 
& Qwen2-VL, Qwen2.5-VL, UI-TARS 
& Divide by $W$ and $H$ \\

Reasoning trace outputs 
& UI-TARS, Qwen2-VL, Qwen2.5-VL, Qwen3-VL 
& Apply think-block stripping before regex parsing \\

\bottomrule

\end{tabular}
}

\vspace{3pt}
\small{\textit{Note: Steps where coordinate extraction fails are recorded as \textbf{No Prediction}.}}

\end{table}

\subsubsection{Evaluation Parameters}

All models are evaluated using a unified set of inference hyperparameters to ensure a fair and reproducible comparison. These parameters are summarized in Table~\ref{tab:parameters}.
A prediction is considered correct if the predicted coordinate lies inside the ground-truth bounding box $B_t$. For \textbf{GUI-Actor}, we additionally allow a $\pm14$ pixel patch overlap tolerance. This tolerance reflects the model’s coordinate-free grounding mechanism, which operates over visual patch tokens of approximately this spatial size.
Certain models occasionally produce malformed outputs that cannot be parsed into valid coordinates. In particular, \textbf{UI-TARS} may emit incomplete responses when decoded greedily. To mitigate spurious failures, we allow a single retry using sampling with temperature $0.3$ when parsing fails. This retry mechanism reduces no-prediction errors while preserving deterministic behavior for successful predictions.

\begin{table}[t]
\centering
\caption{Shared inference hyperparameters used across all evaluated models.}
\label{tab:parameters}

\small
\renewcommand{\arraystretch}{1.25}
\setlength{\tabcolsep}{6pt}

\rowcolors{2}{rowshade}{white}

\resizebox{\textwidth}{!}{
\begin{tabular}{c c c}

\toprule
\rowcolor{tableheader}
\textbf{Parameter} & \textbf{Value} & \textbf{Description} \\
\midrule

Tolerance $\delta$ 
& $14$ px patch overlap 
& Applied only to GUI-Actor; other models require the point to lie inside $B_t$ \\

Top-K predictions 
& $K=5$ (GUI-Actor), $K=1$ (others) 
& GUI-Actor outputs multiple candidate coordinates \\

Retries 
& $2$ (UI-TARS), $1$ (others) 
& UI-TARS retries once with temperature $0.3$ on parse failure \\

Maximum tokens 
& $512$ 
& Applied uniformly across all models \\

\bottomrule
\end{tabular}
}

\end{table}

\section{Statistical Robustness}
\phantomsection
\label{sec:stats}

Although \textbf{MedSPOT} contains 216 tasks, the dataset was intentionally curated to prioritize annotation accuracy and realistic clinical workflows rather than scale alone. To quantify the statistical reliability of our reported metrics under this dataset size, we estimate uncertainty using bootstrapped confidence intervals.

\subsection{Bootstrapped Confidence Intervals}

We compute bootstrapped 95\% confidence intervals for both \textbf{Task Completion Accuracy (TCA)} and \textbf{Step Hit Rate (SHR)} across all 16 evaluated models. Bootstrapping is performed using 1000 resamples with replacement from the set of $n=216$ tasks. The confidence intervals correspond to the 2.5th and 97.5th percentiles of the resulting bootstrap distributions.
The results are summarized in Table~\ref{tab:bootstrap_ci}. The intervals confirm that the key findings of the benchmark remain stable under task sampling variation.

\noindent\textbf{GUI-Actor} is the only model clearly separated from the remaining models, with a TCA confidence interval of $[36.92, 50.00]$ that does not overlap with any other model. In contrast, \textbf{Qwen3-VL} and \textbf{UI-TARS} exhibit overlapping intervals of $[28.97, 41.13]$ and $[24.30, 37.38]$, respectively, indicating that their relative ranking is not statistically significant.
Several models like \textbf{Llama 3.2}, \textbf{Gemma 3}, \textbf{DeepSeek-VL2}, \textbf{GPT-4o-mini}, and \textbf{Qwen2-VL} show TCA intervals of $[0.00, 0.00]$, confirming that they fail to complete any multi-step task regardless of sampling variation.
Finally, \textbf{OS-Atlas} and \textbf{AGUVIS} achieve identical TCA values of $26.64\%$ with heavily overlapping confidence intervals, suggesting that their performance is statistically indistinguishable at the current dataset scale.

\begin{table}[t]
\centering
\caption{Bootstrapped 95\% confidence intervals for TCA and SHR across all models ($n=216$ tasks, 1000 bootstrap resamples).}
\label{tab:bootstrap_ci}

\small
\renewcommand{\arraystretch}{1.2}
\setlength{\tabcolsep}{5pt}

\rowcolors{2}{rowshade}{white}

\resizebox{\textwidth}{!}{
\begin{tabular}{lcccc}

\toprule
\rowcolor{tableheader}
\textbf{Model} & \textbf{TCA (\%)} & \textbf{95\% CI (TCA)} & \textbf{SHR (\%)} & \textbf{95\% CI (SHR)} \\
\midrule

GUI-Actor & 43.46 & [36.92, 50.00] & 49.58 & [42.88, 56.24] \\
Qwen3-VL & 35.05 & [28.97, 41.13] & 46.59 & [40.61, 52.68] \\
UI-TARS & 30.84 & [24.30, 37.38] & 41.93 & [35.89, 48.68] \\
OS-Atlas & 26.64 & [20.56, 32.71] & 55.01 & [48.93, 60.76] \\
AGUVIS & 26.64 & [21.03, 32.71] & 54.76 & [48.75, 60.63] \\
CogAgent & 15.42 & [10.75, 20.56] & 37.80 & [30.48, 44.85] \\
Qwen2.5-VL & 12.62 & [8.41, 17.29] & 19.30 & [14.97, 23.78] \\
GPT-5 & 2.80 & [0.93, 5.14] & 16.53 & [12.13, 21.49] \\
SeeClick & 1.40 & [0.00, 3.27] & 4.33 & [2.40, 6.44] \\
UGround & 0.93 & [0.00, 2.34] & 10.17 & [6.14, 14.23] \\
Qwen2-VL & 0.00 & [0.00, 0.00] & 1.83 & [0.47, 3.60] \\
Mistral-3 & 0.00 & [0.00, 0.00] & 6.14 & [3.17, 9.32] \\
GPT-4o-mini & 0.00 & [0.00, 0.00] & 4.91 & [2.29, 7.42] \\
DeepSeek-VL2 & 0.00 & [0.00, 0.00] & 2.73 & [0.93, 4.89] \\
Gemma 3 & 0.00 & [0.00, 0.00] & 1.33 & [0.51, 2.41] \\
Llama 3.2 & 0.00 & [0.00, 0.00] & 0.00 & [0.00, 0.00] \\

\bottomrule
\end{tabular}
}

\end{table}

\subsection{Per-Software Statistical Note}
Per-software results should be interpreted as indicative trends rather than definitive rankings. Platforms with few tasks. notably Orthanc (10 tasks) and Bluelight (12 tasks), have wide confidence intervals and limited statistical power. The consistent ranking of ITK-SNAP as easiest and RadiAnt 
as hardest across all 16 models provides convergent evidence despite per-platform scale limitations. We plan to expand \textbf{MedSPOT} to 500+ tasks in future work to improve per-software statistical power.

\FloatBarrier
\section{Data Provenance \& Ethics}
\phantomsection
\label{sec:ethics}

\subsection{Data Provenance}
All GUI screenshots were captured from publicly available medical imaging software platforms operating under their default configurations on a standard desktop environment. No real patient data was used at any stage of dataset construction, annotation, or evaluation.
All DICOM images originate from the Pseudo-PHI-DICOM-Data dataset~\cite{pseudophi2021}, released by The Cancer Imaging Archive (TCIA) under a CC BY 4.0 license. This dataset contains de-identified medical images with synthetically generated Protected Health Information (PHI) inserted for the purpose of evaluating medical image de-identification algorithms. Consequently, no real patient identity, clinical history, or protected health information is present in the images. The dataset spans multiple imaging modalities, including CT, MRI, PT, MG, DX, and CR, providing modality diversity representative of clinical imaging workflows.
All bounding box annotations were verified against the corresponding interaction videos to ensure spatial accuracy. Tasks exhibiting ambiguous or inconsistent interface states were removed from the final dataset. Inter-annotator agreement statistics are reported in Appendix~\ref{sec:iaa}.

\subsection{Ethical Considerations}
This benchmark does not involve human subjects, clinical trials, patient recruitment, or personally identifiable data. Therefore, Institutional Review Board (IRB) approval was not required.
\textbf{MedSPOT} is intended strictly for research purposes, specifically for evaluating the spatial grounding capabilities of multimodal large language models (MLLMs) in clinical GUI environments.

\section{Failure Taxonomy Definitions}
\phantomsection
\label{sec:failure}

Errors in GUI grounding can arise from multiple sources, including perceptual resolution limits, spatial imprecision, and semantic misunderstanding. To disentangle these factors, we introduce a structured failure taxonomy that assigns each incorrect prediction to a mutually exclusive category.
Formally, let the model prediction at step $t$ be $\hat{p}_{t}$, or a set of predictions $\hat{P}_{t}$ for multi-point models, and let $\hat{B}_{t}$ denote the ground-truth bounding box of the target UI element. Examples of each failure mode are illustrated in Fig.~\ref{fig:no_prediction}--\ref{fig:farmiss}.

\subsection{No Prediction (NP)}

\begin{figure}[t]
\centering
\includegraphics[width=\linewidth]{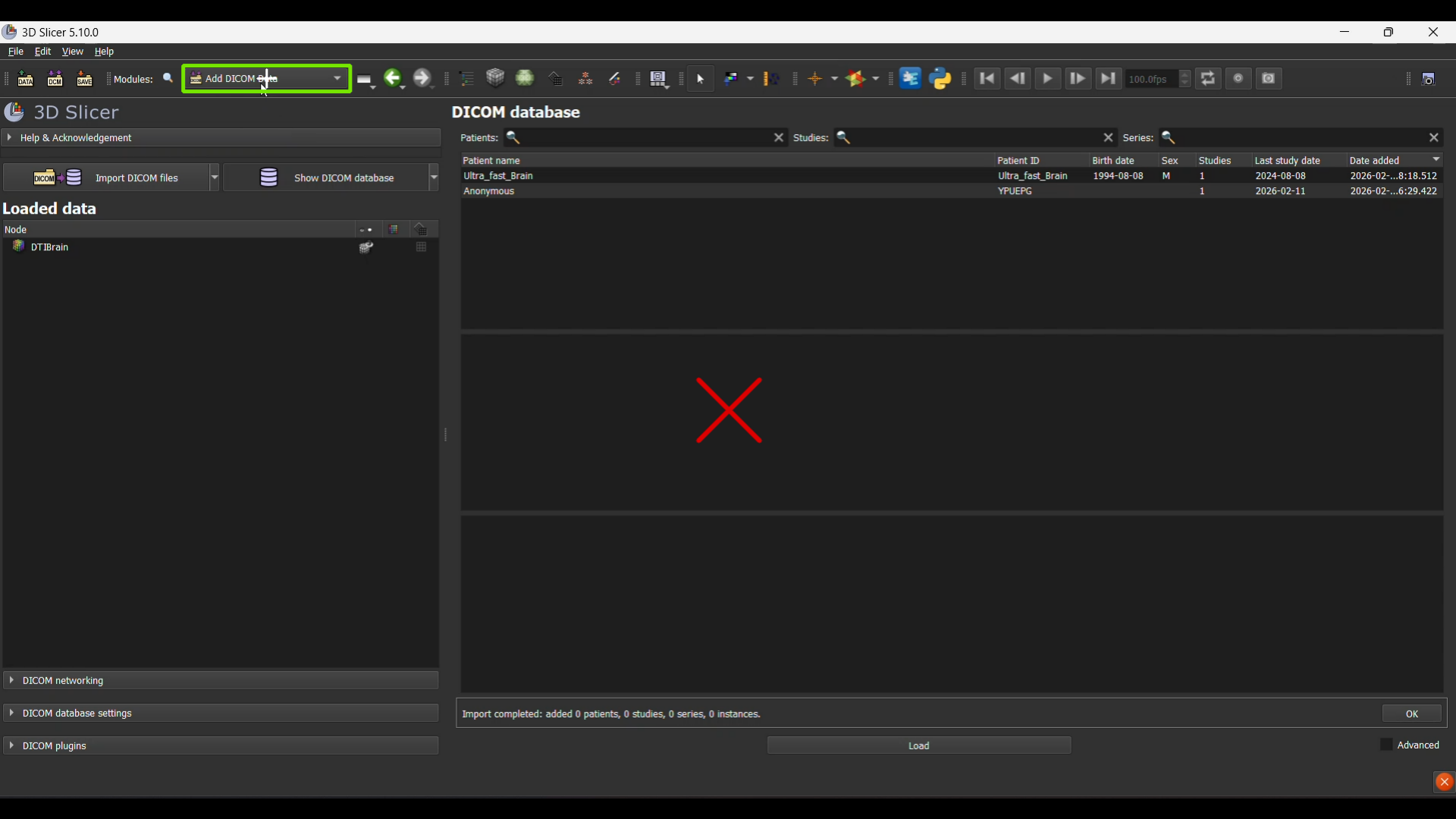}
\caption{Example of a \textbf{No Prediction (NP)} failure. Although the target element \textit{“Add DICOM Data”} is visible in the interface, the model output does not contain a valid coordinate pair, resulting in an empty prediction set $\hat{P}_{t}=\emptyset$. This reflects a decoding or instruction-following failure rather than a spatial localization error.}
\label{fig:no_prediction}
\end{figure}

\textbf{Definition.} If
$
\hat{P}_{t} = \emptyset
$
the step is classified as \textbf{No Prediction (NP)}.
This category captures cases where no valid coordinate pair can be extracted from the model output. Such failures typically arise from malformed outputs, violations of output formatting constraints, or the inability to convert the instruction into an actionable representation. Importantly, NP represents a breakdown in the language-to-action interface rather than a spatial grounding error.
An example of this failure mode is shown in Fig.~\ref{fig:no_prediction}, where the model does not produce a valid coordinate despite the target element being clearly visible in the interface.

\subsection{Small Target (ST)}

\begin{figure}[t]
\centering
\includegraphics[width=\linewidth]{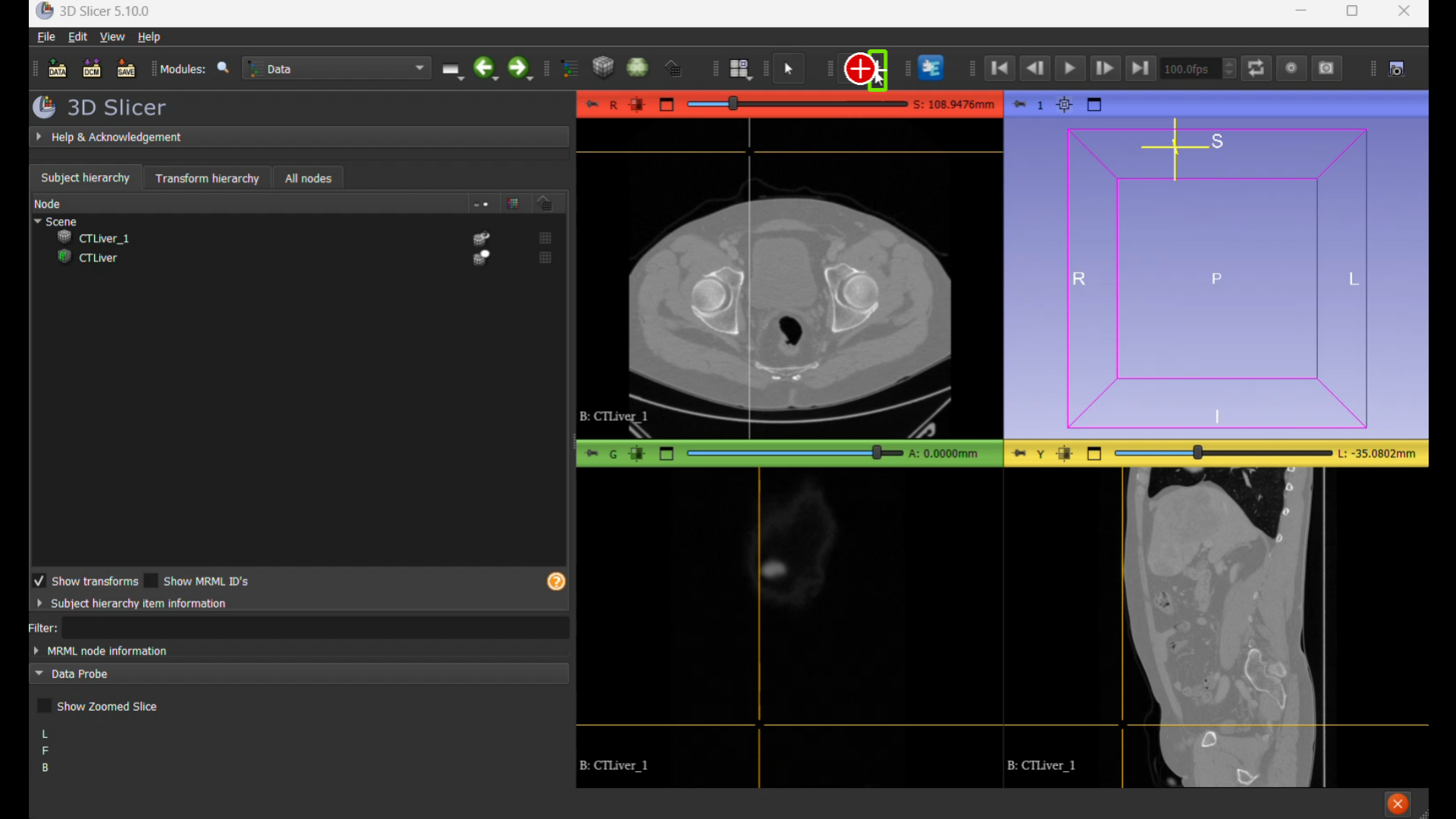}
\caption{Example of a \textbf{Small Target (ST)} failure. The actionable interface element occupies a very small portion of the GUI, requiring extremely precise localization. Even when the instruction is understood, predicted coordinates may fall outside the target due to the resolution limits of the vision encoder.} 
\label{fig:small_target}
\end{figure}

\textbf{Definition.} If the normalized area of the target satisfies
\[
A(B_t) < 4 \times 10^{-4},
\]
the failure is attributed to the \textbf{Small Target (ST)} category.
This category isolates errors caused by extreme spatial precision requirements. When the clickable region occupies only a very small portion of the screen, even semantically correct predictions may fall outside the target.
From a perceptual perspective, modern vision encoders discretize images into patches of size $p \times p$. If

\[
A(B_t) \ll \frac{p^2}{WH},
\]

the target may not be represented as a distinct visual token, making precise localization inherently difficult. Thus, ST failures reflect limitations in visual resolution rather than semantic misunderstanding. An example of this scenario is illustrated in Fig.~\ref{fig:small_target}, where the actionable icon is extremely small relative to the full GUI canvas.

\subsection{Near Miss (NM)}

\textbf{Near Miss (NM)} captures predictions that are geometrically close to the correct element but fail strict containment within the ground-truth bounding box.

We define two complementary criteria:

\begin{itemize}

\item \textbf{Expanded bounding box criterion.}

Let the expanded region be

\[
B'_t = Expand(B_t, \alpha), \quad \alpha > 1.
\]

If

\[
\hat{p}_t \in B'_t \setminus B_t,
\]

the prediction is classified as a near miss.

\item \textbf{Centroid distance criterion.}

Let $c(B_t)$ denote the center of the bounding box. If

\[
\frac{\|\hat{p}_t - c(B_t)\|_2}{\sqrt{W^2 + H^2}} < \tau_n,
\]

with $\tau_n \approx 0.03$, the prediction is also classified as NM.

\end{itemize}

\begin{figure}[t]
\centering
\includegraphics[width=\linewidth]{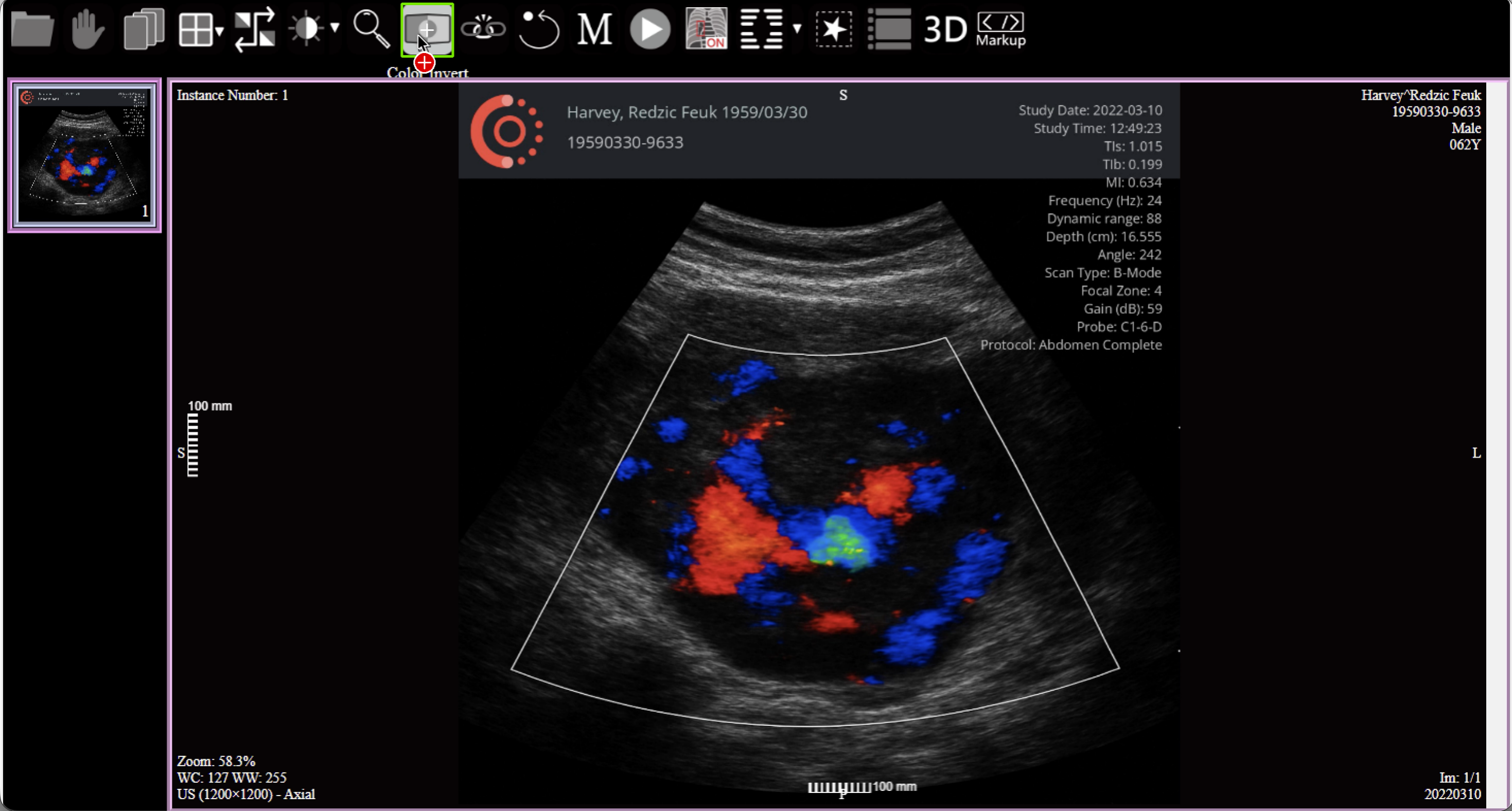}
\caption{Example of a \textbf{Near Miss (NM)} failure. The predicted coordinate lies close to the correct UI element but outside the ground-truth bounding box. This reflects correct semantic grounding with insufficient localization precision.}
\label{fig:nearmiss}
\end{figure}

Near misses indicate correct semantic grounding but insufficient spatial precision. Such errors often arise from coarse visual features, coordinate decoding noise, or ambiguity at pixel-level resolution. A high proportion of near misses suggests that the model is close to solving the task and may benefit from improved spatial refinement. Near-miss failures therefore indicate correct semantic grounding but insufficient spatial precision. As illustrated in Fig.~\ref{fig:nearmiss}, the predicted coordinate lies close to the correct UI element but falls outside the ground-truth bounding box.

\subsection{Edge Bias (EB)}

\textbf{Definition.} A prediction is classified as \textbf{Edge Bias (EB)} if

\[
\hat{x}_t < \tau_e W \quad \vee \quad \hat{x}_t > (1-\tau_e)W
\]

or symmetrically for the vertical coordinate, where $\tau_e = 0.05$.

\begin{figure}[t]
\centering
\includegraphics[width=\linewidth]{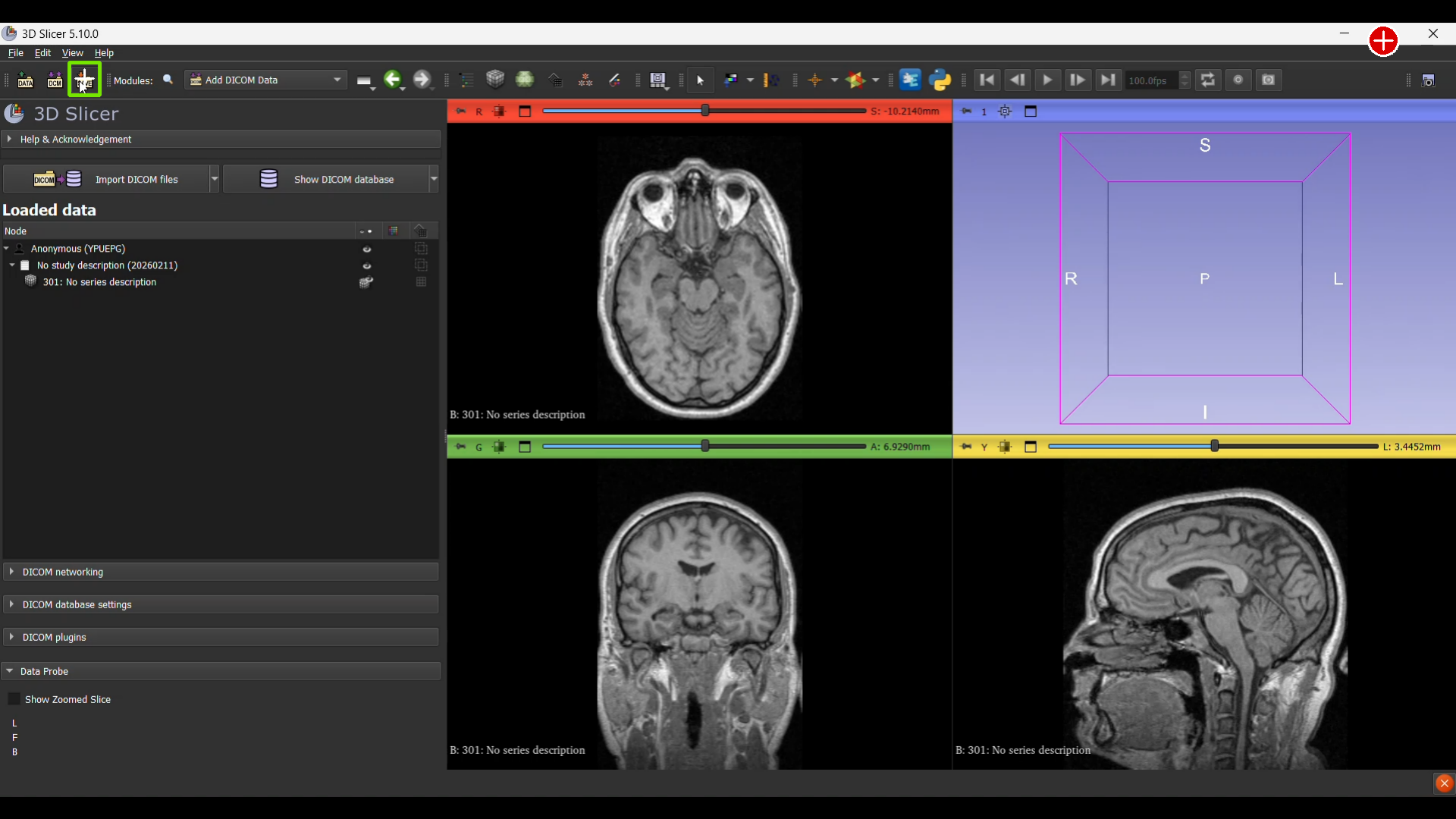}
\caption{Example of an \textbf{Edge Bias (EB)} failure. The model predicts a coordinate near the image boundary rather than the correct interface element, indicating a systematic bias in the spatial prediction distribution.}
\label{fig:edgebias}
\end{figure}

This category captures systematic biases where predictions collapse toward image boundaries. Such behavior may arise from dataset artifacts or prior collapse in spatial attention mechanisms.
From a probabilistic perspective, this indicates that the spatial distribution

\[
p(\hat{p} \mid I,s)
\]

contains spurious probability modes near image borders that are independent of the instruction. An example of this systematic behavior is shown in Fig.~\ref{fig:edgebias}, where the predicted coordinate collapses toward the image boundary rather than the relevant interface element.

\begin{figure}[htbp]
\centering
\includegraphics[width=\linewidth]{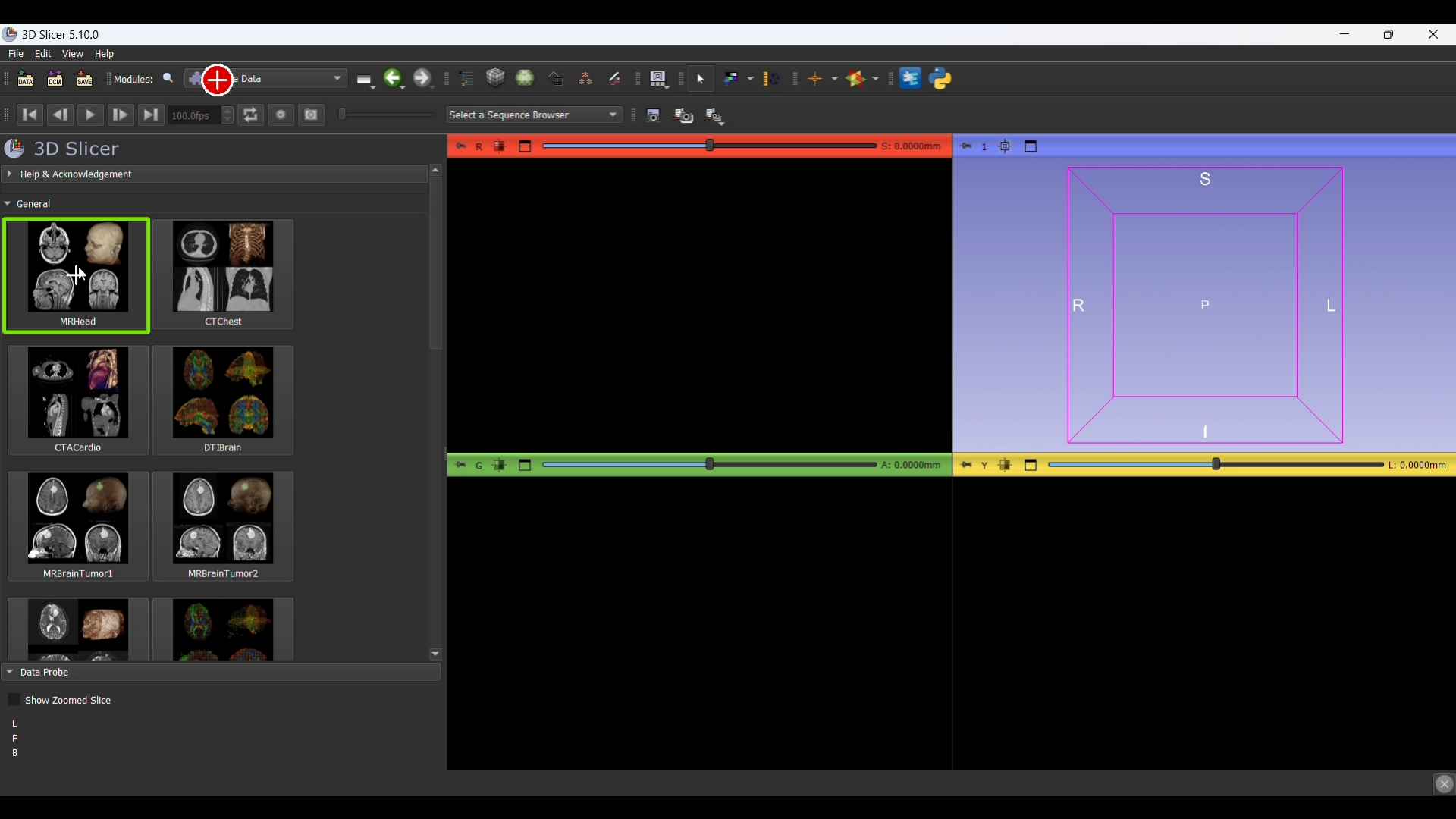}
\caption{Example of a \textbf{Toolbar Confusion (TC)} failure. The predicted coordinate lies in the global toolbar region rather than the task-relevant element within the workspace, indicating a contextual grounding error.}
\label{fig:toolbarconfusion}
\end{figure}
\subsection{Toolbar Confusion (TC)}

\textbf{Definition.} A prediction is classified as \textbf{Toolbar Confusion (TC)} if

\[
\hat{y}_t < \tau_{tb} H, \quad \tau_{tb}=0.12.
\]

This category captures a recurring phenomenon in GUI environments: confusion between persistent global UI components (e.g., menus and toolbars) and task-relevant elements.

The top region of many interfaces contains visually salient controls that appear across tasks. Models often over-attend these areas due to structural repetition or weak grounding between the instruction and localized interface regions.
This phenomenon is illustrated in Fig.~\ref{fig:toolbarconfusion}, where the model selects a control in the global toolbar region instead of the task-relevant element in the workspace.
\begin{figure}[htbp]
\centering
\includegraphics[width=\linewidth]{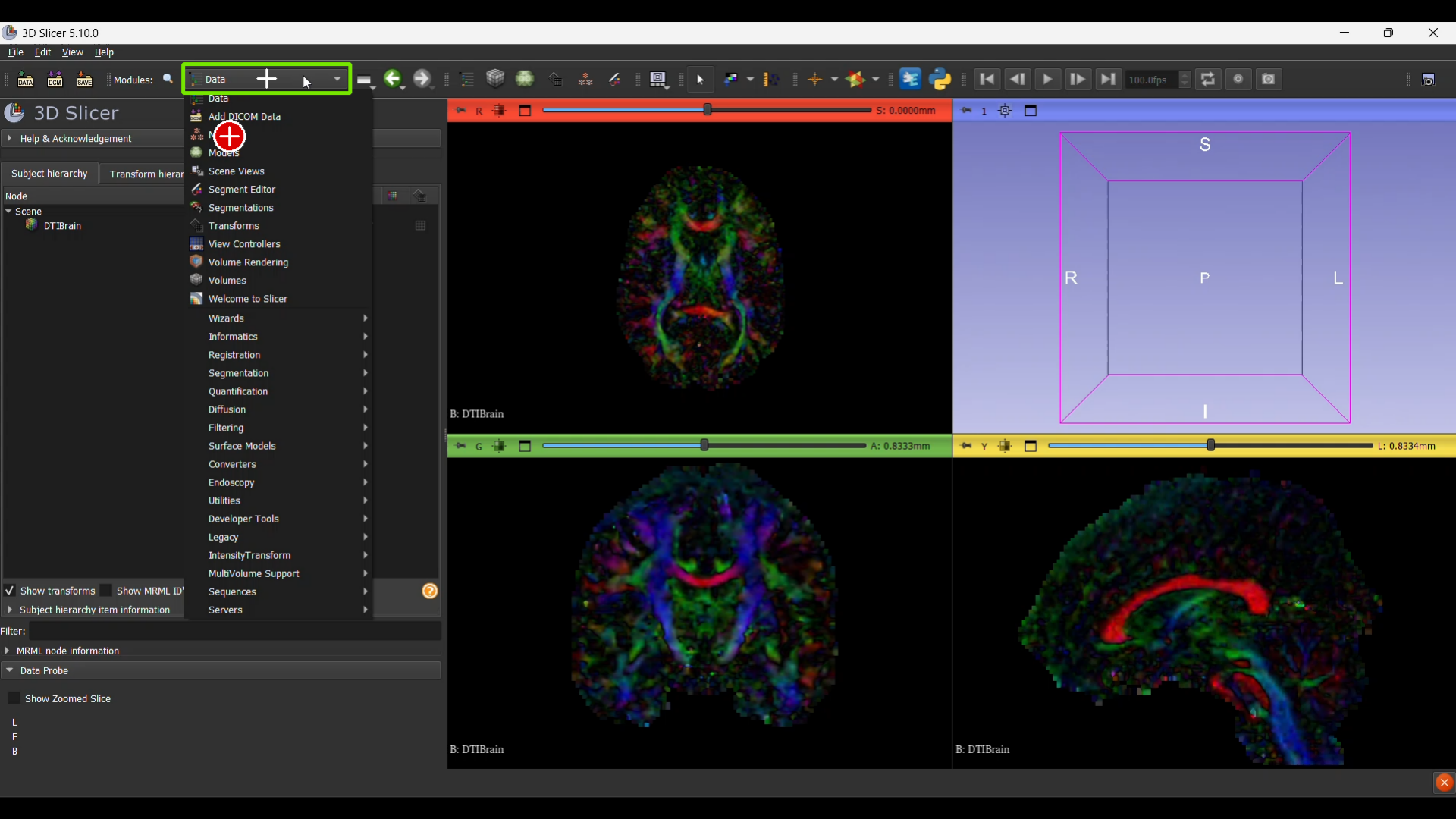}
\caption{Example of a \textbf{Far Miss (FM)} failure. The predicted coordinate is far from the correct UI element and corresponds to an unrelated interface control, indicating a breakdown in instruction understanding and visual grounding.}
\label{fig:farmiss}
\end{figure}

\subsection{Far Miss (FM)}

All remaining incorrect predictions are categorized as \textbf{Far Miss (FM)}.
These predictions lie outside the expanded target region and are not explained by edge bias or toolbar confusion. Far misses represent true semantic grounding failures, where the model fails to align the instruction with the correct UI element.

The taxonomy is applied in a priority-ordered manner, ensuring that each failure is attributed to the most causally informative factor. A representative example is shown in Fig.~\ref{fig:farmiss}, where the predicted coordinate lies far from the correct UI element and corresponds to an unrelated control.

\section{Insights}
\phantomsection
\label{sec:insights}

At each interaction step $t$, the model must predict a precise click location conditioned on the current GUI state and instruction. Formally, the task can be expressed as learning the conditional distribution

\[
\hat{p}_t \sim P(p \mid I_t, s_t, \mathcal{H}_t),
\]

where $I_t$ denotes the GUI screenshot, $s_t$ is the instruction containing latent references to interface elements, $\mathcal{H}_t$ represents the interaction history, and $p$ is the pixel-level action coordinate. This formulation highlights that GUI interaction is a compositional problem involving visual perception, language grounding, spatial reasoning, and temporal consistency.

However, most multimodal large language models (MLLMs) are not trained to directly approximate this distribution. Instead, their training objective typically optimizes

\[
P(\text{text} \mid \text{image}, \text{text}),
\]

where the goal is to generate textual outputs conditioned on multimodal inputs. In contrast, GUI interaction requires predicting a structured spatial action,

\[
P(\text{coordinate} \mid \text{image}, \text{instruction}),
\]

which introduces a fundamental mismatch between the model’s training objective and the requirements of precise spatial grounding. This mismatch manifests in three key limitations.

\subsection{Spatial Quantization}

A fundamental limitation of modern vision–language models arises from the discretization of visual inputs into finite-resolution tokens~\cite{massih2026qvlm}. Most vision encoders used in MLLMs adopt patch-based tokenization, where an input image $I \in \mathbb{R}^{H \times W \times 3}$ is partitioned into non-overlapping patches:

\[
I \rightarrow \{v_1, v_2, \ldots, v_n\}.
\]

Each token $v_i$ corresponds to a spatial region of size $p \times p$ and encodes aggregated information over that patch rather than pixel-level details.

Let the ground-truth bounding box for the target UI element be $B_t$ with area

\[
A(B_t) = (x_2 - x_1)(y_2 - y_1).
\]

If

\[
A(B_t) < p^2,
\]

the target lies below the spatial resolution of a single visual token. In this regime, the target may either be entirely contained within one patch or distributed across patch boundaries without being distinctly represented.

Consequently, the model’s spatial reasoning operates at the patch level:

\[
(i,j) \in \left[1,\frac{H}{p}\right] \times \left[1,\frac{W}{p}\right].
\]

The model can therefore localize the approximate region containing the target but cannot resolve the exact pixel-level location $(x^*, y^*)$ within that patch. This limitation manifests empirically as \textit{small-target} and \textit{near-miss} failures, where the predicted point lies close to the correct element but fails precise containment.

\subsection{Continuous Actions from Discrete Tokens}

A second challenge arises from the mismatch between the continuous action space required by GUI interaction and the discrete token generation mechanism of autoregressive language models~\cite{xu2025structuring,jing2025sparkui}.

The task requires predicting a spatial action represented as a continuous coordinate

\[
p = (x,y) \in \mathbb{R}^2,
\]

implying that an ideal model would approximate a continuous conditional distribution

\[
P(p \mid I,s).
\]

However, MLLMs generate sequences of discrete tokens:

\[
P(w_1, w_2, \ldots, w_n \mid I,s),
\]

where each $w_i$ belongs to a vocabulary $\mathcal{V}$. Producing a coordinate therefore requires implicitly learning a mapping

\[
f : \mathcal{V}^* \rightarrow \mathbb{R}^2.
\]

This indirect representation introduces several sources of instability. First, coordinates must be generated digit by digit, meaning that small token-level errors can produce large spatial deviations:

\[
\Delta x = |x_{\text{pred}} - x_{\text{true}}|.
\]

Second, the mapping from text tokens to coordinates is not uniquely defined. Models may output coordinates in multiple formats (e.g., $(x,y)$, \texttt{x=a, y=b}, or \texttt{[x,y]}), which complicates reliable parsing.

Third, coordinate systems may be ambiguous. Predictions may appear in normalized ranges $(0,1)$, percentages $(0,100)$, or absolute pixels. Without explicit grounding, scale inconsistencies can produce systematic localization errors where

\[
\hat{p}_{\text{scaled}} \notin B_t.
\]

Finally, under uncertainty, autoregressive models tend to generate high-probability token patterns. In spatial contexts, this often results in predictions biased toward boundary values or common coordinate ranges. Empirically, these behaviors manifest as \textit{edge bias} and \textit{toolbar confusion} failures.

Thus, many errors arise not solely from poor reasoning but from a representational mismatch between the continuous action space and discrete token generation.

\subsection{Lack of Explicit Spatial Supervision}

A third limitation stems from the absence of direct spatial supervision during the training of most MLLMs~\cite{qiu2025spatial,tang2025vpp}. In an ideal spatial prediction framework, the model would learn a direct mapping

\[
\hat{p} = f_\theta(I,s),
\]

where $\hat{p} \in \mathbb{R}^2$ represents the predicted coordinate. Such a model could be trained with a regression objective

\[
\mathcal{L}_{\text{spatial}} = \|\hat{p} - p^*\|_2^2,
\]

where $p^*$ denotes the ground-truth location. This formulation provides direct geometric supervision and penalizes spatial errors explicitly.

In contrast, MLLMs are typically trained using an autoregressive language objective:

\[
\mathcal{L}_{\text{LLM}} = -\sum_i \log P(w_i \mid w_{<i}, I, s),
\]

where $w_i$ are tokens in the output sequence. As a result, the model learns

\[
P(\text{text} \mid I,s)
\]

rather than

\[
P(p \mid I,s).
\]

Spatial grounding therefore emerges indirectly through cross-modal alignment learned from image–caption datasets, instruction-following corpora, and multimodal conversations~\cite{chen2024lion}. Because these datasets rarely provide pixel-level supervision, the mapping

\[
(I,s) \rightarrow p
\]

is learned implicitly rather than explicitly optimized. Consequently, spatial reasoning remains an emergent capability rather than a directly trained objective.

\section{Limitations: Visual Examples}
\phantomsection
\label{sec:limitations_appendix}

This section provides visual illustrations of the primary limitations discussed in the main paper. Each example highlights practical scenarios where current evaluation assumptions may diverge from real-world clinical GUI environments.

\subsection{Benchmark Scale}

\begin{figure}[t]
\centering
\begin{minipage}{0.48\linewidth}
\centering
\includegraphics[width=\linewidth]{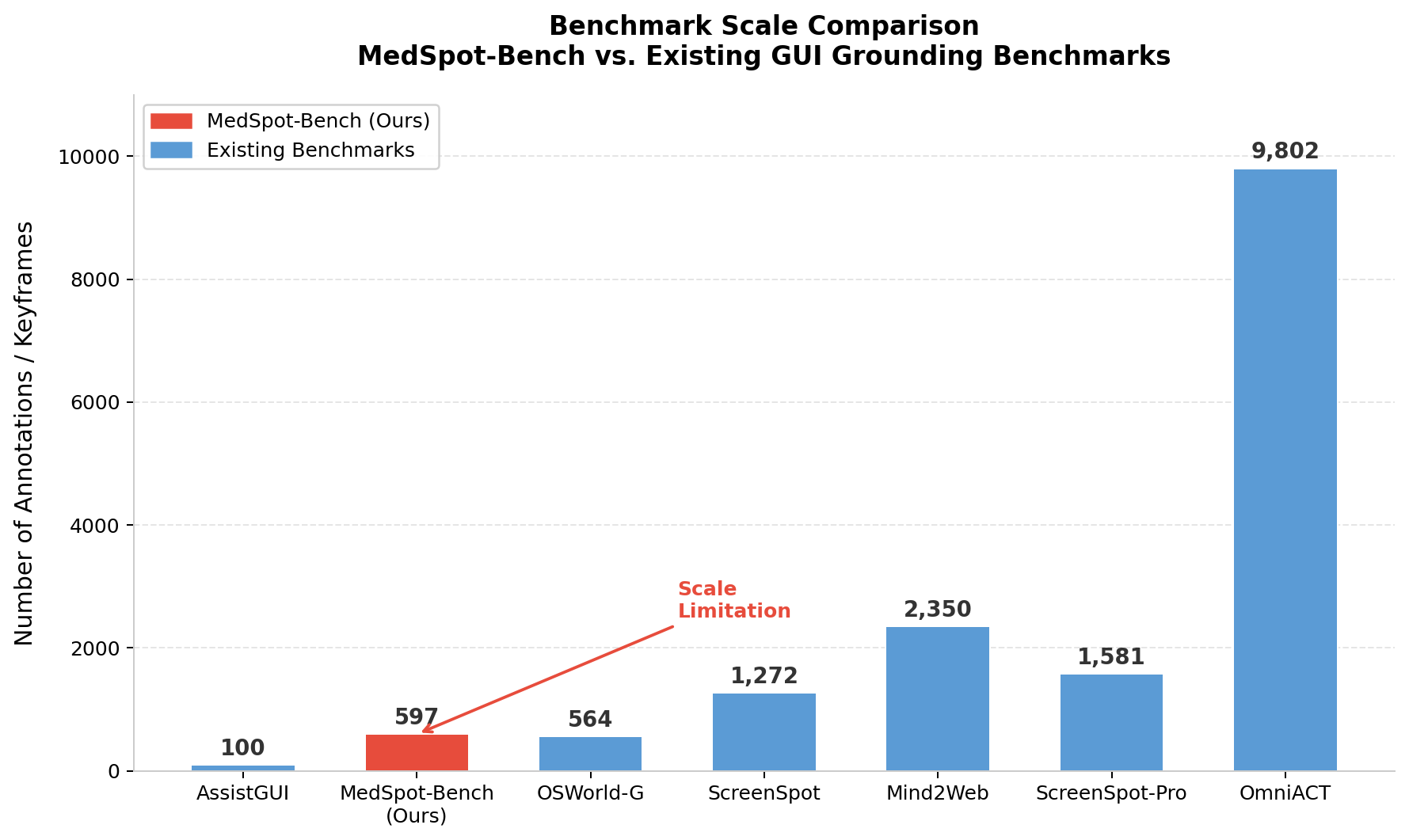}
\end{minipage}
\hfill
\begin{minipage}{0.48\linewidth}
\centering
\includegraphics[width=\linewidth]{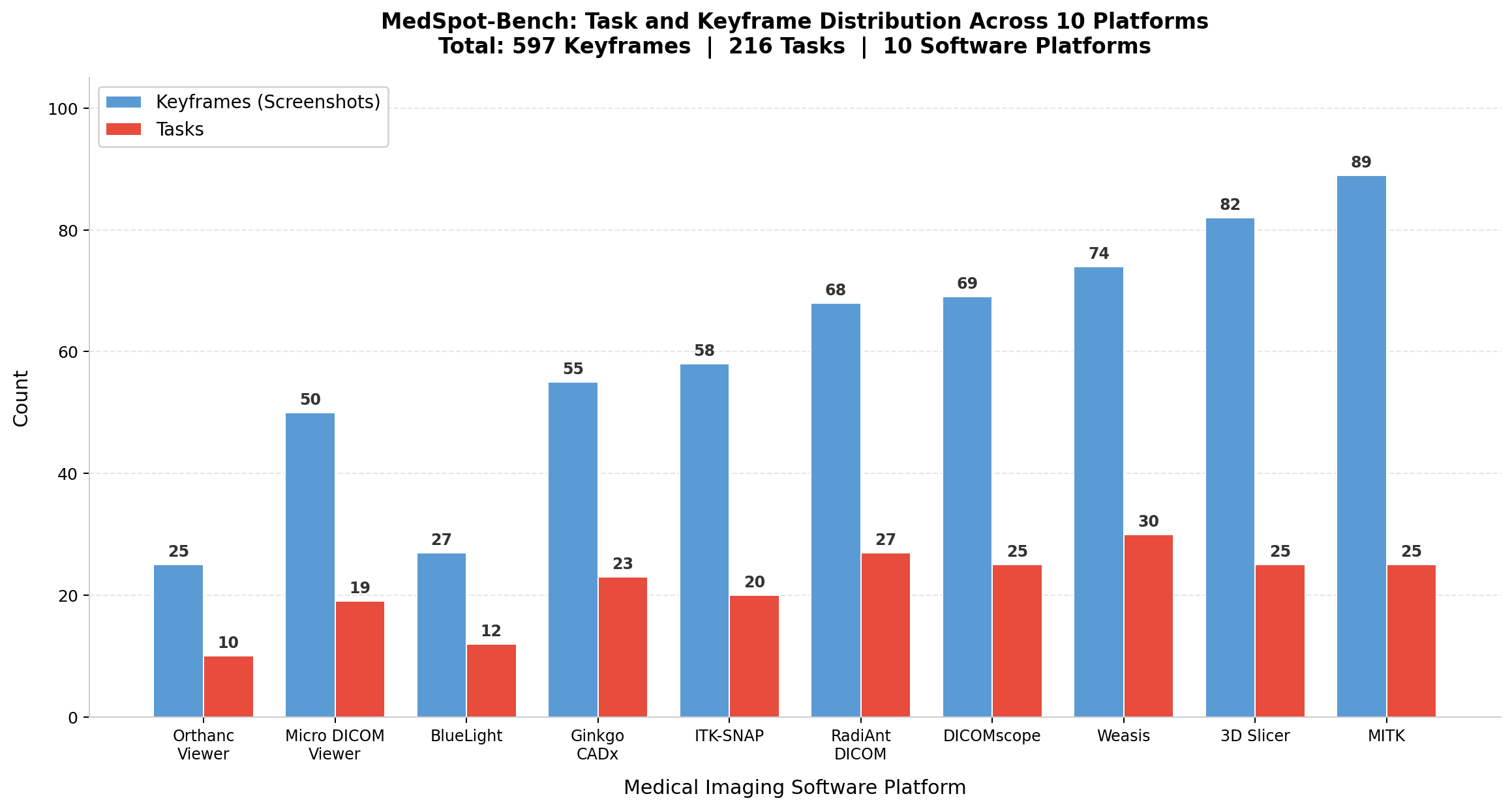}
\end{minipage}
\caption{Scale and task distribution of \textbf{MedSPOT}.}
\label{fig:limit_scale}
\end{figure}
\begin{figure}[t]
\centering
\includegraphics[width=0.75\linewidth]{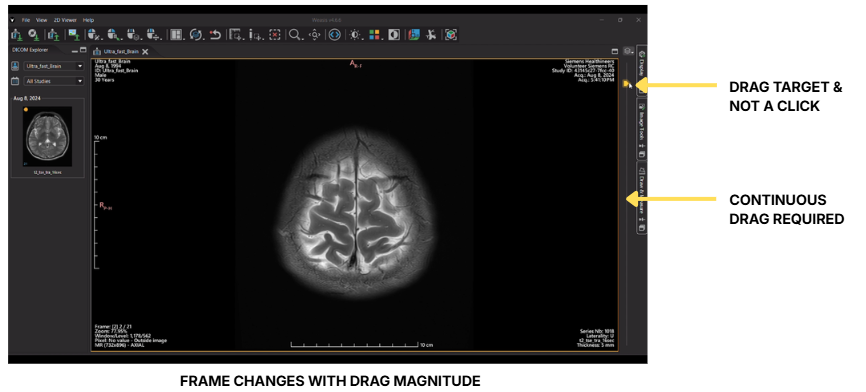}
\caption{Example of a slider-based interaction requiring drag input.}
\label{fig:limit_drag}
\end{figure}
As shown in Fig.~\ref{fig:limit_scale}, \textbf{MedSPOT} contains 597 annotated keyframes spanning 216 multi-step tasks. While this scale is smaller than many general-purpose GUI benchmarks, the dataset prioritizes annotation quality and workflow realism over volume. Each task requires careful reconstruction of clinically meaningful workflows within specialized medical imaging platforms.
Unlike general-purpose benchmarks where tasks can be crowdsourced at scale~\cite{deng2023mind2web,kapoor2024omniact}, constructing tasks in clinical software requires domain-aware annotation and validation across multiple platforms. Expanding the benchmark to include additional workflows, platforms, and interaction scenarios remains an important direction for future work.

\subsection{Click-Only Interactions}

Currently, \textbf{MedSPOT} focuses exclusively on click-based interactions. However, many real-world GUI operations involve more complex interaction types such as drag, scroll, or keyboard input. 
Figure~\ref{fig:limit_drag} illustrates an example of a DICOM frame navigation slider that requires a drag operation to move between frames. Such interactions are common in medical imaging software but are not yet represented in the benchmark, limiting the scope of interaction behaviors that can be evaluated.

\subsection{Multi-Tool Ambiguity}

\begin{figure}[t]
\centering
\includegraphics[width=\linewidth]{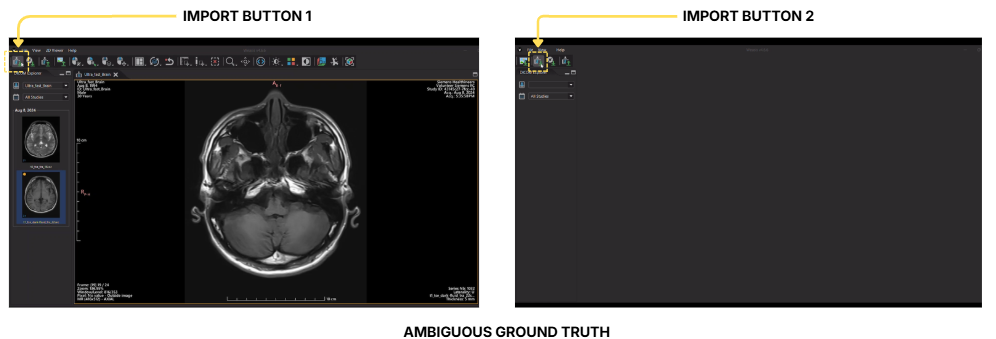}
\caption{Example of multiple visually distinct controls leading to the same operation.}
\label{fig:limit_ambiguity}
\end{figure}

Some medical interfaces expose multiple controls that trigger the same functionality. As illustrated in Fig.~\ref{fig:limit_ambiguity}, several buttons may lead to the same destination within the application. 
This creates ambiguity when defining a single ground-truth click target, since multiple interface elements could be considered correct from a functional perspective. Current annotations select one canonical target, but future work could explore evaluation strategies that support multiple valid interaction targets.

\subsection{Sequential Evaluation Early Termination}

\begin{figure}[t]
\centering
\includegraphics[width=\linewidth]{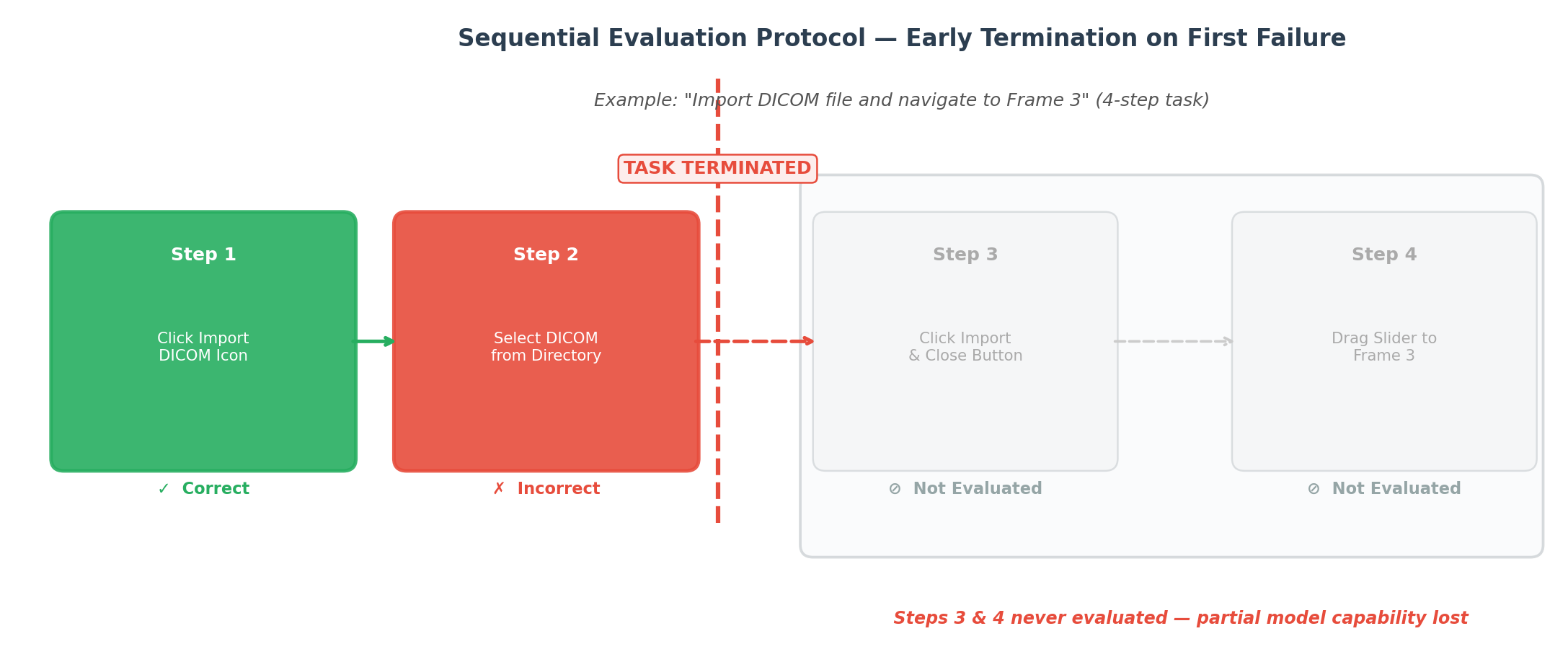}
\caption{Example illustrating early termination in sequential evaluation.}
\label{fig:limit_termination}
\end{figure}
\begin{figure}[htbp]
\centering
\includegraphics[width=\linewidth]{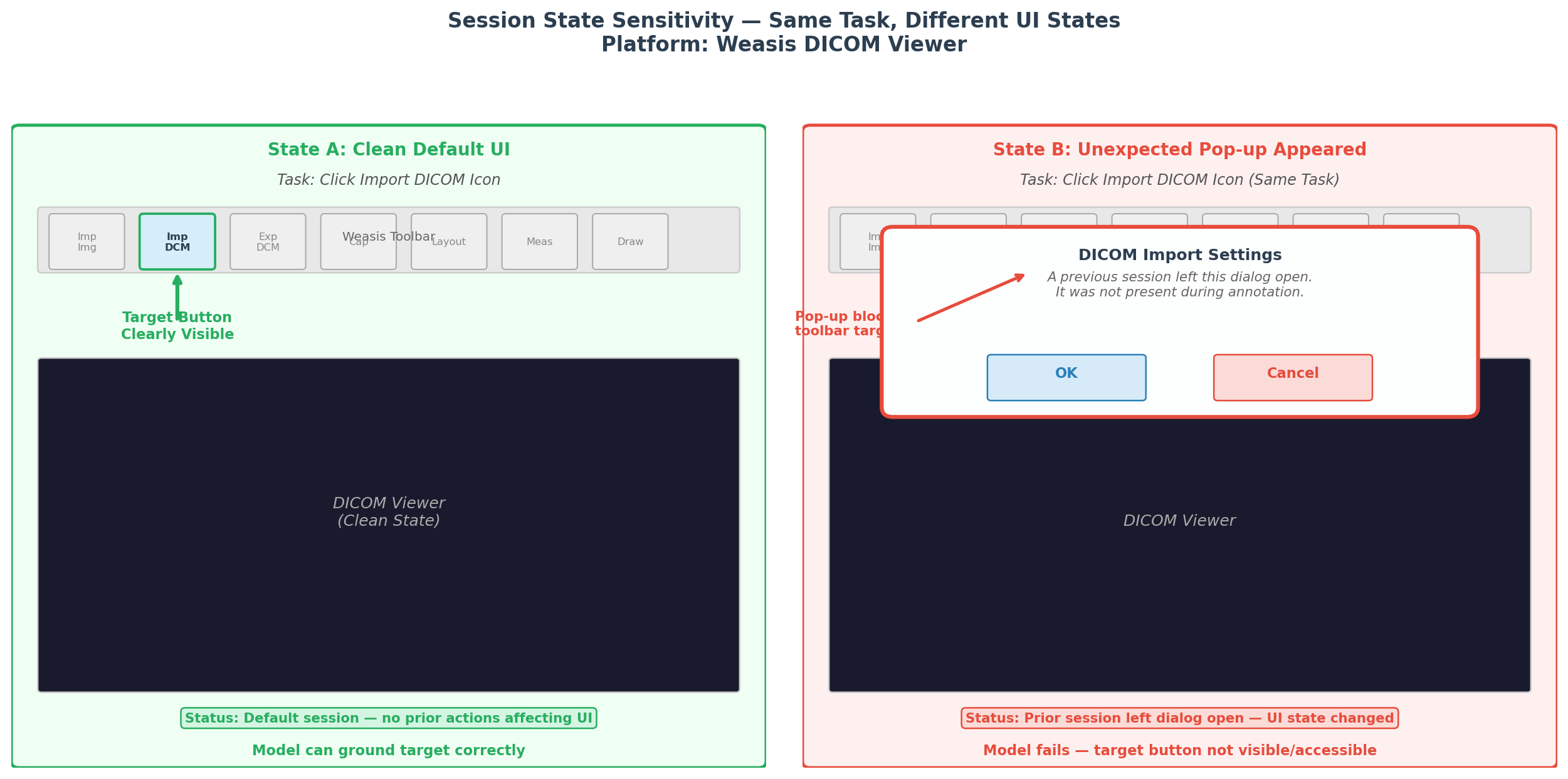}
\caption{Example of session-state changes affecting GUI layout and accessibility of target elements.}
\label{fig:limit_session}
\end{figure}
\textbf{MedSPOT} follows a strict sequential evaluation protocol in which task scoring terminates immediately after the first incorrect grounding step. As shown in Fig.~\ref{fig:limit_termination}, a model may correctly complete the first step but fail at the second, preventing subsequent steps from being evaluated.
This protocol follows established sequential task evaluation methodologies~\cite{xie2024osworld}, where later actions are considered unevaluable once a prerequisite step fails. However, this approach can underestimate partial task-level capability, since models may successfully perform later steps even after an early mistake.

\subsection{Session State Sensitivity}

All screenshots in \textbf{MedSPOT} were captured under a fixed default software configuration. However, real-world medical software usage is highly sensitive to session state, where prior user actions can significantly alter the interface layout.
As illustrated in Fig.~\ref{fig:limit_session}, a dialog window left open from a previous interaction may overlay important interface elements. In the example shown, a DICOM Import Settings dialog blocks the toolbar region containing the target button that is visible in the clean default state.
Such dynamic UI changes—including floating panels, loading dialogs, or rearranged layouts—are not represented in the static keyframe dataset. As a result, models evaluated on clean screenshots may perform correctly in the benchmark but fail under altered session states encountered in real clinical workflows.

\subsection{Synthetic DICOM Data and Risk-Unaware Metrics}

\textbf{MedSPOT} relies exclusively on synthetic or publicly available demo DICOM datasets across all evaluated platforms~\cite{pseudophi2021}. This ensures compliance with privacy and ethical guidelines, but synthetic data lacks the variability, noise, artifacts, and anatomical diversity present in real clinical imaging environments.

Furthermore, the benchmark currently employs risk-unaware evaluation metrics. All incorrect clicks are penalized equally, regardless of their potential clinical impact. For example, selecting the wrong patient record during report generation carries far greater consequences than clicking an incorrect toolbar icon, yet both are treated identically during evaluation.
Future work should explore risk-aware evaluation frameworks that assign higher penalties to errors affecting clinically critical actions.

\begin{figure}[t]
\centering
\includegraphics[width=\linewidth]{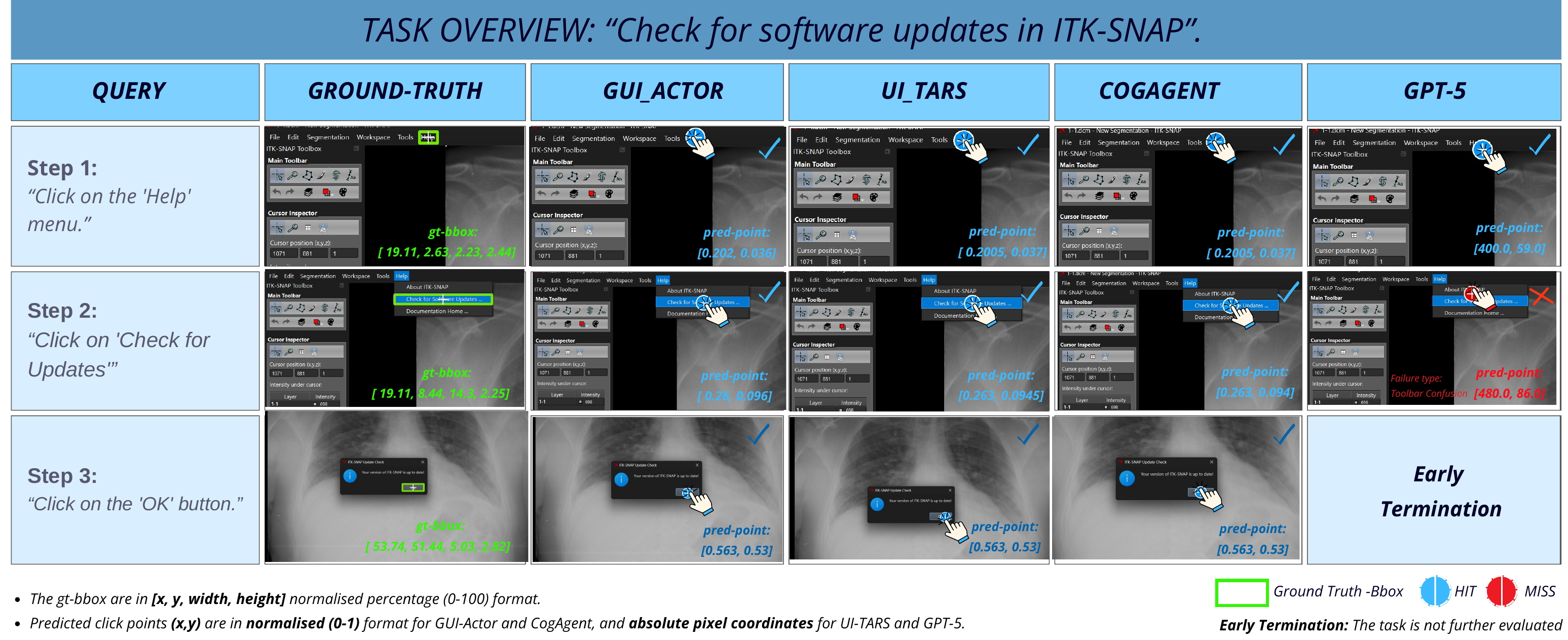}
\caption{\textbf{Easy Interface (ITK-SNAP): ``Check for software updates''.} 
The interface contains a small number of clearly separated toolbar elements. GUI-Actor, UI-TARS, and CogAgent correctly ground all three steps of the task, while GPT-5 fails only at Step~2. The minimal visual clutter and clear spatial separation of controls facilitate reliable grounding.}
\label{fig:qaeasy}
\end{figure}
\section{Qualitative Analysis: Visual Examples}
\phantomsection
\label{sec:qualitative_appendix}

This section provides qualitative examples illustrating how different interface layouts affect model performance. The examples highlight how interface simplicity, visual density, and semantic ambiguity influence spatial grounding behavior across models.

\subsection{Easy Interface Example}

Figure~\ref{fig:qaeasy} shows a representative task from the \textit{ITK-SNAP} interface where the instruction requires checking for software updates. This interface features a clean layout with a minimal number of toolbar icons and clearly separated UI elements.
In this scenario, three of the evaluated models, \textbf{GUI-Actor}, \textbf{UI-TARS}, and \textbf{CogAgent} successfully complete all sequential steps required by the task. Their predictions remain consistently aligned with the intended interface elements throughout the interaction sequence. 
In contrast, \textbf{GPT-5} fails at Step~2 despite correctly identifying the first element. This suggests that even when models initially ground the instruction correctly, maintaining spatial consistency across multiple sequential steps remains challenging. 
Overall, the example illustrates that uncluttered interfaces with large, visually distinct controls significantly reduce spatial ambiguity and enable more reliable grounding performance across models.
\begin{figure}[t]
\centering
\includegraphics[width=\linewidth]{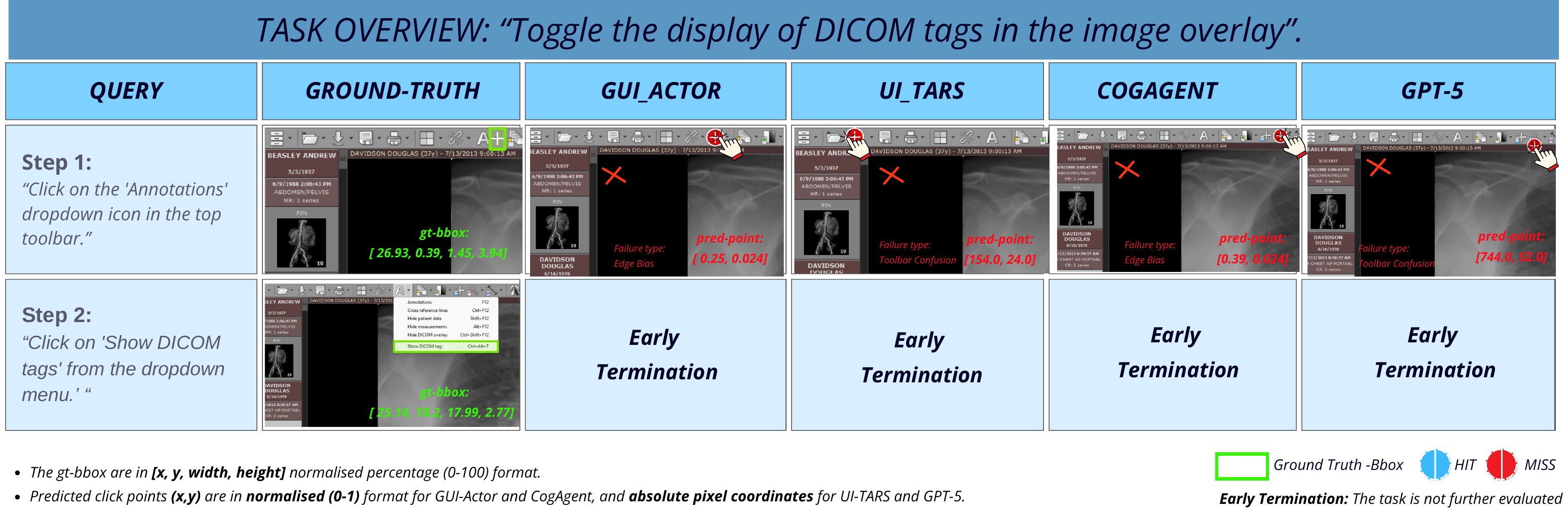}
\caption{\textbf{Hard Interface (RadiAnt): ``Toggle DICOM tags in image overlay''.} 
The dense toolbar layout causes all models to fail at Step~1. Predictions collapse toward the top boundary of the interface, producing edge-bias and toolbar-confusion failures.}
\label{fig:qahard}
\end{figure}

\subsection{Hard Interface Example}

In contrast, Fig.~\ref{fig:qahard} presents a task from the \textit{RadiAnt} interface involving the instruction to toggle DICOM tags in the image overlay. This interface contains a densely populated toolbar with numerous visually similar icons located along the top boundary of the screen.
All four evaluated models fail at the very first step of the task. The predictions cluster along the upper edge of the interface, producing failure modes classified as \textit{Edge Bias} and \textit{Toolbar Confusion}. Instead of identifying the correct icon, the models are attracted to visually salient but semantically irrelevant controls in the toolbar region.
Because the sequential evaluation protocol terminates upon the first incorrect step, none of the models proceed to Step~2, resulting in a task completion accuracy of zero for this task across all models. This example demonstrates how dense interface layouts and visually repetitive controls substantially increase the difficulty of spatial grounding.

\subsection{Unsolved Task Example}

\begin{figure}[h]
\centering
\includegraphics[width=\linewidth]{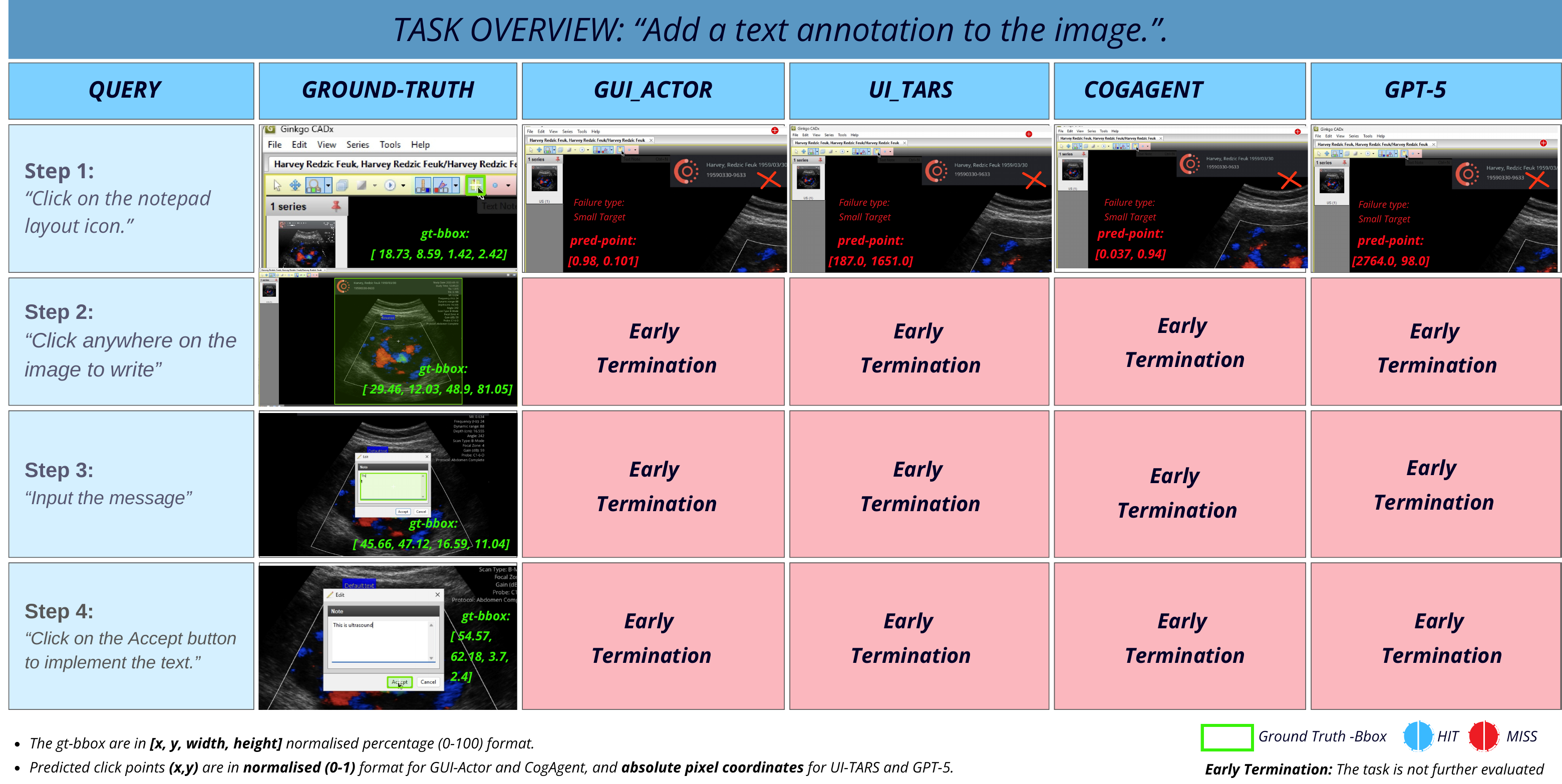}
\caption{\textbf{Unsolved Task (Ginkgo CADx): ``Add a text annotation to the image''.} 
All evaluated models fail at Step~1. The target element is a small icon embedded within a dense toolbar, making it both visually ambiguous and difficult to localize precisely.}
\label{fig:allfail}
\end{figure}

Figure~\ref{fig:allfail} illustrates a particularly challenging task from the \textit{Ginkgo CADx} interface requiring the addition of a text annotation to the image. In this case, all four evaluated models fail at the first interaction step.
This task belongs to a subset of 40 tasks within \textbf{MedSPOT} where none of the evaluated models achieve a correct grounding step. The target element corresponds to a small notepad-style icon embedded within a dense toolbar containing many visually similar symbols.

Two factors make this task particularly difficult. First, the icon itself occupies a very small spatial region relative to the interface, requiring extremely precise localization. Second, the icon’s visual appearance does not have a clear semantic correspondence with the natural language instruction “add text annotation.” As a result, models must rely on weak visual cues rather than direct semantic alignment.
This example highlights the limitations of current multimodal models when operating in specialized professional interfaces that contain domain-specific icons and tightly packed controls.

\newpage
\section{Inter-Annotator Agreement}
\phantomsection
\label{sec:iaa}

To validate annotation quality, a randomly sampled overlap subset of the benchmark was independently annotated by all three annotators, who were kept blind to each other's annotations throughout the process.Because annotators provided free-form textual descriptions of the target GUI elements rather than selecting from a fixed label vocabulary,
we assess annotation reliability based on the spatial agreement of bounding boxes, which uniquely identify the GUI element associated with each instruction.

\begin{table}[h]
\centering
\caption{Pairwise Bounding Box Inter-Annotator Agreement.
Mean IoU and GIoU are reported with 95\% bootstrap confidence
intervals (1,000 resamples, seed = 42).
Agreement Rate (AR) is the proportion of frames where
IoU exceeds the stated threshold.}
\label{tab:iaa_bbox}

\small
\setlength{\tabcolsep}{6pt}
\renewcommand{\arraystretch}{1.2}

\rowcolors{2}{rowshade}{white}

\begin{tabular}{lcccccc}
\toprule
\rowcolor{tableheader}
\color{white}\multirow{1}{*}{\textbf{Pair}} &
\multicolumn{2}{c}{\color{white}\textbf{Mean IoU}} &
\multicolumn{2}{c}{\color{white}\textbf{Mean GIoU}} &
\multicolumn{2}{c}{\color{white}\textbf{AR (\%)}} \\

\rowcolor{tableheader}
& \color{white}Value & \color{white}95\% CI 
& \color{white}Value & \color{white}95\% CI 
& \color{white}$\geq$0.50 & \color{white}$\geq$0.75 \\
\midrule

A1 vs.\ A2 & 0.909 & [0.899, 0.919] & 0.898 & [0.887, 0.910] & 100.0 & 99.3 \\
A1 vs.\ A3 & 0.842 & [0.799, 0.885] & 0.832 & [0.790, 0.874] & 100.0 & 97.8 \\
A2 vs.\ A3 & 0.869 & [0.858, 0.880] & 0.864 & [0.852, 0.876] & 100.0 & 98.6 \\

\midrule

\rowcolor{green!12}
\textbf{Overall} &
\textbf{\color{highlight}0.874} &
\textbf{[0.840, 0.899]} &
\textbf{\color{highlight}0.865} &
\textbf{[0.829, 0.892]} &
\textbf{100.0} &
\textbf{98.6} \\

\bottomrule
\end{tabular}

\end{table}

\paragraph{Spatial Agreement.}
Bounding-box agreement was quantified using the Intersection over
Union (IoU) and Generalised IoU (GIoU)~\cite{rezatofighi2019giou},
with GIoU providing additional penalties for non-overlapping
predictions. As shown in Table~\ref{tab:iaa_bbox}, all annotator
pairs achieved strong spatial agreement, with mean IoU values
exceeding $0.84$. The closest pair (A1 vs.\ A2) achieved
$\text{IoU} = 0.909$ (95\% CI: [0.899, 0.919]).
Across all annotator pairs, the overall mean IoU was
$\mathbf{0.874}$ (95\% CI: [0.840, 0.899]), indicating highly
consistent spatial localisation of GUI elements.
Agreement rates were near perfect, with all frames exceeding
$\text{IoU} \geq 0.50$ and more than $98\%$ of frames exceeding
$\text{IoU} \geq 0.75$. These results demonstrate that annotators
consistently identified the same interface element with high
spatial precision.

\paragraph{Reliability Assessment.}
The high IoU and GIoU agreement values indicate that annotators consistently localised the same GUI elements across independent annotations. These results demonstrate that the annotation protocol yields reliable and reproducible spatial labels, supporting the use of the dataset as a benchmark for GUI grounding in medical software interfaces.
\end{document}